\documentclass[runningheads]{llncs}

\usepackage{eccv}

\usepackage{eccvabbrv}

\usepackage{graphicx}
\usepackage{booktabs}
\usepackage{tabularx}
\usepackage{multirow}
\usepackage{amsmath}
\usepackage{array}
\newcolumntype{Y}{>{\centering\arraybackslash}X}
\usepackage{pifont}

\usepackage[accsupp]{axessibility}  

\newif\ifdrafting
\draftingtrue 

\ifdrafting
    
    \newcommand{\mc}[1]{\textcolor{magenta}{[Matthew: #1]}}
    \newcommand{\sd}[1]{\textcolor{blue}{[Shalini: #1]}}
    \newcommand{\js}[1]{\textcolor{green}{[Josef: #1]}}
    \newcommand{\kn}[1]{\textcolor{red}{[Koki: #1]}}
    \newcommand{\xt}[1]{\textcolor{orange}{[Xueting: #1]}}
    
\else
    \newcommand{\mc}[1]{}
    \newcommand{\sd}[1]{}
    \newcommand{\js}[1]{}
    \newcommand{\kn}[1]{}
    \newcommand{\xt}[1]{}
\fi

\usepackage[hyperfootnotes=false]{hyperref}

\usepackage{orcidlink}

\begin{document}

\title{Coherent 3D Portrait Video Reconstruction via Triplane Fusion} 

\author{Shengze Wang\href{https://mcmvmc.github.io/}{\inst{1}}\inst{,2*} \and
Xueting Li\href{https://sunshineatnoon.github.io/}{\inst{2}} \and
Chao Liu\href{https://research.nvidia.com/person/chao-liu}{\inst{2}} \and
Matthew Chan\href{https://matthew-a-chan.github.io/}{\inst{2}} \and
Michael Stengel\href{https://research.nvidia.com/person/michael-stengel}{\inst{2}} \and
Josef Spjut\href{https://josef.spjut.me/}{\inst{2}} \and
Henry Fuchs\href{https://henryfuchs.web.unc.edu/}{\inst{1}} \and
Shalini De Mello\href{https://research.nvidia.com/person/shalini-de-mello}{\inst{2}} \and
Koki Nagano\href{https://luminohope.org/}{\inst{2}}}

\authorrunning{S.~Wang et al.}

\institute{University of North Carolina at Chapel Hill \and
NVIDIA
}

\maketitle

\begin{figure*}[ht]
\centering
\includegraphics[width=\textwidth]{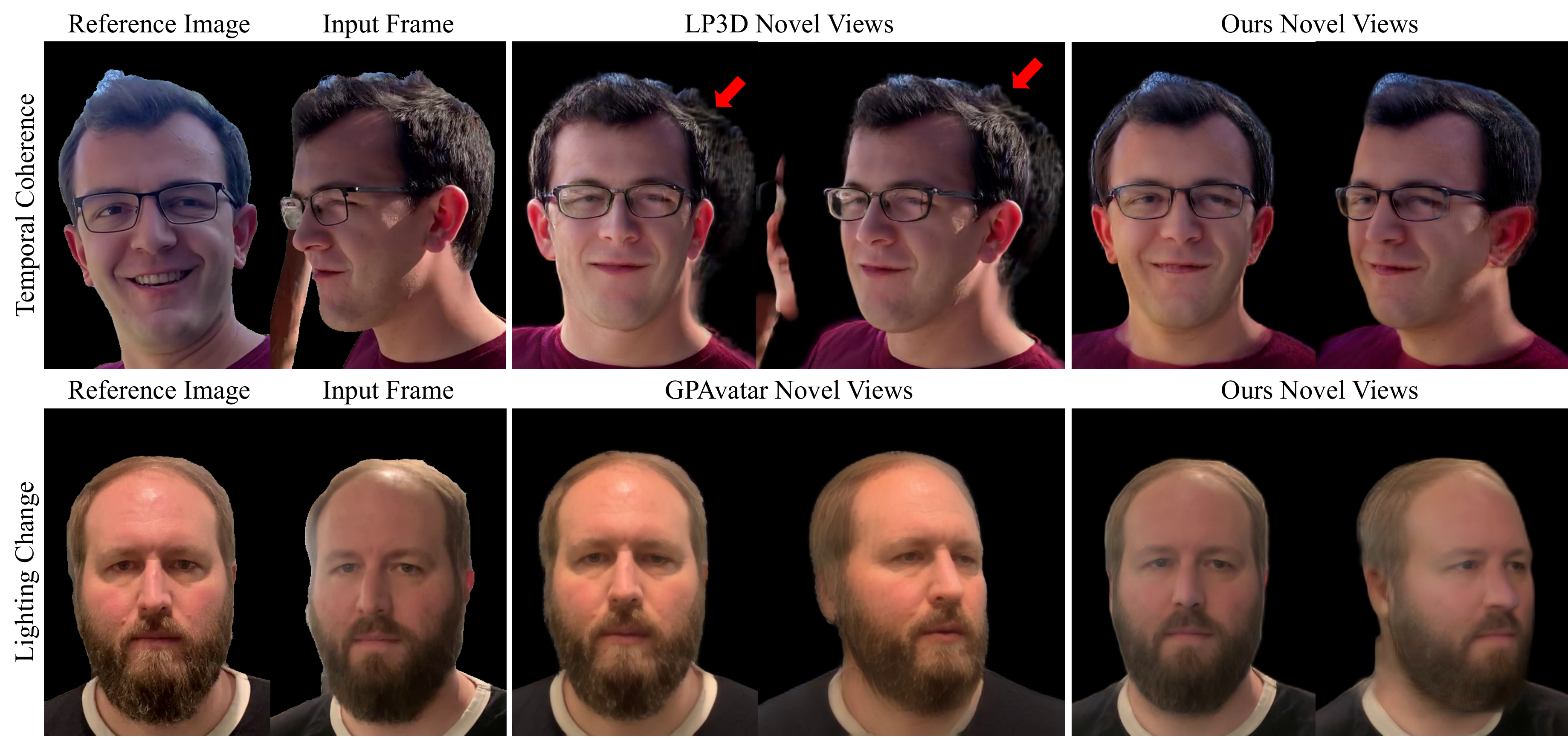}
\caption{We propose a triplane fusion method for reconstructing coherent 3D portrait videos. Our method captures the authentic dynamic appearance of the user (\eg, facial expressions and lighting) while producing temporally coherent 3D videos. Trained only using a synthetic 3D video dataset, our encoder-based method achieves both state-of-the-art 3D reconstruction accuracy and temporal consistency.}
\label{fig:teaser}
\end{figure*}

\let\oldthefootnote\thefootnote
\let\thefootnote\relax\footnote{$^*$This work was done during an internship at NVIDIA}
\let\thefootnote\oldthefootnote

\begin{abstract}
    Recent breakthroughs in single-image 3D portrait reconstruction have enabled telepresence systems to stream 3D portrait videos from a single camera in real-time, potentially democratizing telepresence. 
    However, per-frame 3D reconstruction exhibits temporal inconsistency and forgets the user's appearance. 
    On the other hand, self-reenactment methods can render coherent 3D portraits by driving a personalized 3D prior, but fail to faithfully reconstruct the user's per-frame appearance (\eg, facial expressions and lighting). 
    In this work, we recognize the need to maintain both coherent identity and dynamic per-frame appearance to enable the best possible realism. 
    To this end, we propose a new fusion-based method that fuses a personalized 3D subject prior with per-frame information, producing temporally stable 3D videos with faithful reconstruction of the user's per-frame appearances. 
    Trained only using synthetic data produced by an expression-conditioned 3D GAN, our encoder-based method achieves both state-of-the-art 3D reconstruction accuracy and temporal consistency on in-studio and in-the-wild datasets. \url{https://research.nvidia.com/labs/amri/projects/stable3d}
\end{abstract}

\vspace{-5mm}
\section{Introduction}
\label{sec:intro}
\vspace{-2mm}
Telepresence for bringing distant people face-to-face in 3D, stands out as a particularly compelling application of computer vision and graphics, which can transform human experiences.
Over the last several decades, various successful telepresence systems~\cite{raskar1998office, Kauff2002, PixelCodecAvatar, starline, orts-escolano2016holoportation, MAIMONE2012791, jones2009, Kauff2002} have been developed. However, most employ bulky multi-view 3D scanners or depth sensors to ensure high-quality volumetric per-frame reconstruction. Unlike these classical 3D/4D reconstruction methods, recent AI-based feed-forward 3D lifting techniques, such as LP3D~\cite{trevithick2023} and TriPlaneNet~\cite{bhattarai2024triplanenet}, can lift a single RGB image from an off-the-shelf webcam into a neural radiance field (NeRF) representation in real-time, and pave the path forward towards making 3D telepresence accessible to anyone~\cite{stengel20233dvc}.

Besides democratizing 3D human telepresence, single-frame-based lifting techniques such as LP3D~\cite{trevithick2023}, have the further advantage of faithfully preserving the instantaneous dynamic conditions present in an input video, \eg of lighting, expressions, and posture, all of which are crucial to an authentic telepresence experience. However, single-image reconstruction methods that operate independently on each frame, are not ideal for maintaining temporal consistency. This difficulty stems from the inherent ill-posed nature of single-image-based reconstruction. In order to render novel views that are significantly far from the input view, the system cannot rely on information present in the input view and hence must hallucinate plausible content, which cannot be guaranteed to be consistent across multiple temporal frames (e.g., first row in Fig.~\ref{fig:teaser}, where the frontal view prediction by LP3D~\cite{trevithick2023}  does not match the frontal reference image.). This makes the system susceptible to changes in the lifted 3D portrait's appearance, depending on the user's head pose in the input frame. We find that the most structurally reliable triplane is often produced by an input image with a nearly frontal head pose as shown in Fig.~\ref{fig:distortion} (first column). 

On the opposite spectrum, to single-image-based lifting techniques for telepresence, are 3D self-reenactment methods~\cite{tran2023voodoo, chu2024gpavatar, li2023hidenerf,ye2024real3dportrait}. 
3D reenactment method creates a canonical frame from a reference image (s) representing the appearance of the user (usually a mouth closed, frontal neutral frame; the first column in Fig.~\ref{fig:distortion}), and use a separate driving video to control the facial expressions and poses of the avatar.
They produce temporally consistent results, but do not faithfully reconstruct the input video's dynamic conditions, \ie the actual appearance of the user at the moment such as person-specific expressions, and lighting. It is also not always possible to capture these conditions from the driving video (\eg from cameras in a headset). 
Moreover, reenactment methods often struggle to authentically reconstruct the accurate expressions of the user because the expression control is not precise enough.
Additionally, reenactment methods often fail to reconstruct details not present in the reference image (\eg lighting variation in the bottom row of Fig.~\ref{fig:teaser}; teeth and tongue in Fig.~\ref{fig:comparison}). All of these factors sacrifice realism in 3D reenactment-based methods, making them not ideal for 3D portrait video reconstruction.

In this work, we firstly, recognize the need to maintain both temporal consistency while preserving real-time dynamics of input videos in human telepresence applications. We further address both problems together for the first time in single-view 3D portrait synthesis to enable the best user experience. 
Our key insight to solving this problem is to employ a fusion-based approach to achieve all of these properties: 
the approach needs to leverage the stability and accuracy of a personalized 3D prior, and needs to fuse the prior with per-frame observations to capture the diverse deviations from the prior.

Our model first uses pretrained LP3D~\cite{trevithick2023} to construct a personal triplane prior from a (near) frontal image of the user, which can be casually or passively captured or extracted from a video.  
During video reconstruction, our model uses LP3D to lift each input frame into a raw triplane, which is then fused with the personal triplane prior. When the head pose of the input image is oblique, artifacts and identity distortions may be present in its lifted triplane (see Fig.~\ref{fig:teaser} and \ref{fig:distortion}). Hence we first propose an undistorter module, which learns to undistort the raw instantaneous triplane to more closely match the structure of the correctly-structured prior triplane. We then propose a fuser module, which learns to densely align the undistorted raw triplane to the reference triplane and then fuse the two in a manner that incorporates personalized details such as tattoos or birthmarks present in the reference triplane,  while preserving dynamic lighting, expression and posture information from the input raw triplane.

Similar to LP3D, we also leverage a pre-trained 3D GAN (\ie Next3D\cite{sun2023next3d}, which is an expression-conditioned 3D GAN) as the generator to synthesize dynamic 3D portraits to train our system and thus circumvent the scarcity of real-world 3D portrait data.
We train our model using multiview images rendered from these dynamic 3D portraits. 
Additionally, we perform various augmentations during data generation to enhance the synthetic data such that our model not only learns from expression changes synthesized by Next3D, but also shoulder rotation and different lighting conditions that cannot be synthesized by the pre-trained generator. 

Lastly, we find that the current established evaluation framework~\cite{trevithick2023} only measures a model's ability to reproduce its input image, and relies on visual comparisons to assess the quality of novel views. Since no prior work addresses temporal consistency, their evaluation protocol overlooks this dimension. It is an important consideration in telepresence, where the quality of novel views are often more important than reproducing the input image. To address this gap, we additionally formulate a new evaluation protocol specifically designed to measure a method's robustness to input image's head pose variations as well as the consistency across its novel views.

Major contributions of our work include:
\vspace{-3mm}
\begin{itemize}
    \item We recognize a novel problem: the need to achieve both temporal consistency and reconstruction of dynamic appearances when using single-view 3D lifting solution for telepresence, essential for enabling enhanced user experiences. 
    \item We propose a novel triplane fusion method that fuses the dynamic information from per-frame tripanes with a personal triplane prior extracted from a reference image. 
    Trained only using a synthetic multi-view video dataset, our feedforward approach generates 3D portrait videos that demonstrate both temporal consistency and faithful reconstruction of dynamic appearances (\eg lighting and expression) of the user at the moment, whereas prior solutions can only achieve one of the two properties. 
    \item We propose a new framework to evaluate single-view 3D portrait reconstruction methods using multi-view data. 
    This new framework not only provides accurate evaluation of a method's reconstruction quality by using different viewpoints for evaluation, but also provides insights to a method's robustness by using different viewpoints as inputs.
    \item Evaluations on both in-studio and in-the-wild datasets demonstrate that our method achieves state-of-the-art performance in both temporal consistency and reconstruction accuracy. 
\end{itemize}

\vspace{-3mm}
\section{Related Work}
\vspace{-2mm}
\noindent\textbf{Multi-view 3D/4D face reconstruction.}
Creating high-fidelity 3D or 4D representations of human heads has a long history in computer vision and graphics.
Most earlier works rely on complex multi-view camera systems to capture 3D geometry of faces using multiview stereo algorithms~\cite{Beeler2010,Furukawa2010}, facial performance~\cite{Furukawa2009Dense, Beeler2011,Bradley2010,Bickel2007,Fyffe2015, Wu2018Incremental,Fyffe2017,DynamicFusion} and photorealistic appearance using active illumination systems~\cite{Ghosh2011, Ma2007,Klaudiny2012}. However, they require expensive capture hardware with offline processing and do not scale to the end users for the purpose of generating photorealistic avatars from commodity devices. 

\noindent\textbf{2D portrait reenactment.}
Given a single or a few reference portrait images and a driving video, recent talking-head generators can reenact 2D portraits by transferring the facial expressions and poses from the driving video onto facial portraits~\cite{wang2021facevid2vid, Siarohin_2019_NeurIPS,Zakharov_2019_ICCV,Zakharov20,zhang2022metaportrait,doukas2020headgan,Drobyshev22MP,hong2022depth,tps2022,wang2022latent,StyleHeat2022}. As a 2D portrait generation method, while they can manipulate head poses and expressions of the avatars within 2D portraits to some extent, they do not predict volumetric 3D representations and cannot be rendered from novel view point, which is crucial for 3D telepresence.

\noindent\textbf{3D-aware portrait generation and reenactment.}
To incorporate 3D consistency, some methods combine mesh-based 3D face representations using 3D morphable models (3DMM)~\cite{BlanzVetter1999,FLAME:SiggraphAsia2017} with 2D neural rendering~\cite{DNR2019,wang2023styleavatar,Khakhulin2022ROME,nagano2018pagan,deng2020disentangled,tewari2020stylerig} to embed facial expressions and pose controls in a 2D neural renderer (see the full survey in ~\cite{Tewari2020Neural}). 
Some more recent works use deformable volumetric implicit radiance field-based representations~\cite{mildenhall2020nerf,park2021nerfies,pumarola2020d} or gaussian splatting-based representations~\cite{kerbl3Dgaussians} combined by 
3DMM-based face representations to reconstruct a photorealistic and animatable volumetric head avatar~\cite{Gafni_2021_CVPR,cao2022authentic,zheng2022imavatar,athar2022rignerf,xu2023gaussianheadavatar,qian2023gaussianavatars,saito2024rgca}. However, they tend to require extensive data captures from videos or from a multiview camera setup and person-specific training. Another very recent family of methods use a large-scale video dataset and learn a disentangled triplane 3D representations~\cite{eg3d2022} that allows 3D-aware facial reenactment in a feedforward fashion~\cite{ye2024real3dportrait, chu2024gpavatar,tran2023voodoo, Li2023Oneshot,ma2023otavatar, li2023hidenerf,NOFA2023Yu}. As a reenactment method, these methods construct a canonical 3D representation from a reference image (often a neutral frame), and use facial expressions and head poses extracted from a separate driving video to drive the 3D representation. As such, fine-grained facial expressions may not be captured due to errors in disentanglement. Most importantly, these reenactment methods hallucinate person-specific dynamic appearance of the user (e.g., person-specific wrinkles) since they are not observed in the reference frame.

\noindent\textbf{3D GAN inversion.} 
By combining generative adversarial networks (GAN)~\cite{goodfellow2014generative} and neural volume rendering~\cite{mildenhall2020nerf}, recent breakthroughs in 3D-aware GANs~\cite{eg3d2022,orel2022stylesdf,gu2021stylenerf,deng2022gram,xiang2022gramhd,epigraf,Zhou2021CIPS3D,zhang2022mvcgan,xu2021volumegan,rebain2022lolnerf,xu2022pv3d} demonstrated the unsupervised learning of photorealistic 3D-aware human heads from in the wild 2D images. Notably, EG3D presented triplane 3D representations~\cite{eg3d2022} which can generate photorealistic 3D portraits in real-time. Next3D~\cite{sun2023next3d} extends EG3D to create 3D portrait videos controlled by 3DMM facial expression and pose parameters. We use Next3D to create our synthetic multiview video training data to supervise our network. 
Once these 3D head priors are trained, they can be used to perform single-view 3D reconstruction using GAN inversion to lift a portrait to 3D~\cite{ko20233d3dganinversion,sun2022ide,lin20223dganinversion,Fruehstueck2023VIVE3D}, manipulate the 3D avatar~\cite{hong2021headnerf,sun2022ide,zhuang2022mofanerf}, or 3D personalization~\cite{buehler2023preface, qi2023my3dgen}. Since the single-view 3D GAN inversion is both time consuming and fragile if the camera pose is not optimized together~\cite{ko20233d3dganinversion}, some recent works ~\cite{trevithick2023,bhattarai2024triplanenet} proposed an encoder-based solution. In this work, we build our method on the state-of-the-art single-view triplane encoder LP3D~\cite{trevithick2023}. While it has demonstrated an excellent capability to lift a challenging real-world image to 3D, it lifts every frame independently from scratch, and exhibits temporal inconsistency---a key limitation to create a practical 3D telepresence system. We propose a triplane-fusion-based method which improves the temporal stability of LP3D.
\section{Definitions}
Given that this work targets a new task, we define terminologies in this section to avoid confusion.
We define an "\textbf{input viewpoint}" to be the viewpoint of the input video relative to the user's head.
We use both the terms "\textbf{dynamic appearance}" and "\textbf{per-frame information}" to be the dynamically varying information in a portrait video, such as expressions, lighting condition, and shoulder pose.
We define the "\textbf{input frame}" to be the current video frame that is being converted into a 3D portrait, and the "\textbf{reference image}" to be the image used to capture the prior knowledge of a person, such as a near frontal image. 
We define the triplane reconstructed from the reference image as the "\textbf{triplane prior}" because it encodes a personalized geometric prior about the user.
Frontal images capture both sides of the user's face and thus can help with reconstruction when the input frame captures the user from the sides. 
We redefine these two terms to emphasize that the fusion process is different from self-reenactment: the goal of fusion is to enhance per-frame reconstruction methods with prior information while capturing the authentic dynamic appearance of the user in a video at the same time. 
Dynamic appearance is critical to reconstructing the liveliness of an actual person, whereas reenactment methods focus on driving an avatar instead of reconstructing authentic dynamic appearances in a video.

\vspace{-3mm}
\section{Method}
\begin{figure*}
  \centering
    \includegraphics[width=\textwidth]{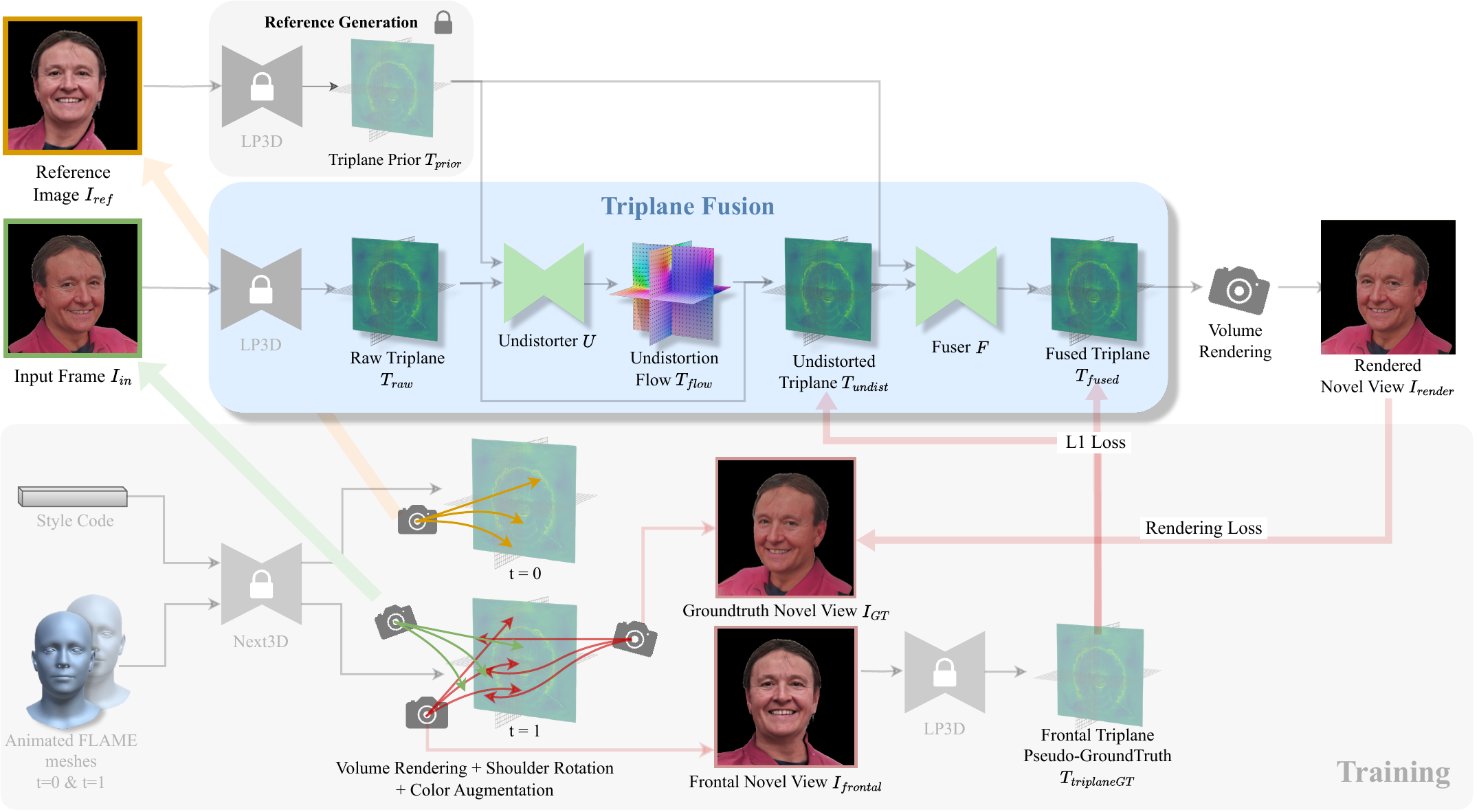}
   \caption{\textbf{Overview}. Given a (near) frontal reference image and an input frame, we reconstruct a triplane prior and a raw triplane respectively using an improved LP3D~\cite{trevithick2023} (Sec.~\ref{sec:lp3d}). Next, we combine these two triplanes through a Triplane Fusion module (blue box) that ensures temporal consistency while capturing realtime dynamic conditions like lighting and shoulder pose (Sec.~\ref{sec:undistorter} and Sec.~\ref{sec:fuser}). Our model is trained with only synthetic video data generated by a 3D GAN~\cite{sun2023next3d}, with carefully designed augmentation methods to account for shoulder motion and lighting changes (Sec.~\ref{sec:data}).
   }
   \label{fig:overview}
   \vspace{-0.5cm}
\end{figure*}

\vspace{-2mm}
Our method aims to reconstruct coherent 3D portrait videos from a monocular RGB video without test-time optimization. 
To improve temporal consistency and reconstruction of occluded areas, we leverage an additional reference image, which can be obtained from the same video or a selfie capture. 
An overview of our method is illustrated in Fig. \ref{fig:overview}. 
Our model first converts an input frame into a raw triplane 
using a frozen pre-trained LP3D (Sec.~\ref{sec:lp3d}). 
Then, the Triplane Undistorter (Sec. \ref{sec:undistorter}) removes view-dependent distortions and artifacts by leveraging the triplane prior, resulting in an undistorted triplane. 
Finally, the Triplane Fuser (Sec. \ref{sec:fuser}) constructs the final triplane by enhancing the undistorted triplane with additional information from the triplane prior.

\subsection{Background: 3D Portrait from a Single Image}
\label{sec:lp3d}
\indent \textbf{LP3D.} The recently proposed LP3D~\cite{trevithick2023} method performs photorealistic 3D portrait reconstruction from a single RGB image. 
Specifically, it uses a feedforward encoder to convert an RGB image into a triplane $\textbf{T} \in \mathbb{R}^{3\times 32\times256\times256}$. 
Then, LP3D performs volume rendering to decode the triplane into an RGB image.
During volume rendering, point samples $\textbf{x}\in \mathbb{R}^3$ are generated by ray marching from the rendering camera. 
These point samples are projected onto each of the three planes (\ie xy-plane, xz-plane, and yz-plane), producing bilinearly interpolated features $f_{xy}$, $f_{xz}$, and $f_{yz}$. 
The three features are averaged to produce the mean feature $f'$, which is then decoded by a lightweight Multi-Layer Perceptron (MLP) into RGB color $\textbf{c}$, density $\sigma$, and a feature vector $\textbf{f}$:
\begin{equation}
    \vspace{-1mm}
    (\textbf{f},\textbf{c},\sigma) = MLP(f').
    \vspace{-1mm}
\end{equation}
These point samples are aggregated together to form pixels through volume rendering~\cite{mildenhall2020nerf}, resulting in a low-resolution feature image $I_f \in \mathbb{R}^{32\times128\times128}$. 
The first 3 channels of $I_f$ are trained to produce an RGB image $I_{low} \in \mathbb{R}^{3\times128\times128}$.
Finally, a lightweight 2D convolutional neural network super-resolves $I_f$ into the final rendering:
\begin{equation}
    \vspace{-1mm}
    SuperRes(I_f) = I_{high} \in \mathbb{R}^{3\times512\times512}.
    \vspace{-1mm}
\end{equation}
LP3D leverages synthetic data generated from a pre-trained 3D GAN (\ie EG3D\cite{eg3d2022}) and thus circumvents the challenging problem of large-scale 3D groundtruth data acquisition. 
During training, LP3D is supervised by both the groundtruth EG3D triplanes as well as the rendered 2D images. 
As a result of the effectively infinite amount of 2D and 3D groundtruth, LP3D is able to generate photorealistic 3D portraits. 
Since it directly maps from images to triplanes via an encoder, LP3D can run in real-time and was thus developed into a complete realtime telepresence system~\cite{stengel20233dvc}. 

\noindent\textbf{Modifications.}
Our implementation of LP3D is slightly different and improved from the original in that we enlarge the cropping of the input portrait image to include more of the shoulders, because shoulders are important for conveying body language and the sense of realism in telepresence. 
Our implementation also includes an additional camera estimator that takes in LP3D's intermediate features to estimate a set of camera parameters $M \in \mathbb{R}^{25}$. 
$M$ represents an estimation of the intrinsic and extrinsic parameters of the camera used to capture the input image. 
The camera estimator allows LP3D to better recreate the input image and improves robustness to inaccurate head poses estimated by off-the-shelf trackers. 
We use our own implementation of LP3D for all evaluations. 

\subsection{Generating Synthetic Dynamic Multiview Data}
\label{sec:data}
Inspired by LP3D~\cite{trevithick2023} and the development of animatable 3D GANs, we use Next3D~\cite{sun2023next3d} to generate animated 3D portraits as groundtruth training data. 
During data preparation, we first pre-process the FFHQ~\cite{karras2019style} dataset into facial landmarks and FLAME~\cite{FLAME:SiggraphAsia2017} coefficients using DECA~\cite{feng2021learning}. 
During the training of our model, we randomly sample a pair of FLAME coefficients and landmarks from the pre-processed dataset and a single style code corresponding to a random identity. 
As shown on the lower left of Fig.~\ref{fig:overview}, these data are input to Next3D to generate a pair of triplanes for $t=0$ and $t=1$, each depicting a different expression of the same synthetic person. 
The triplane for $t=0$ is used to render the reference image from a frontal viewpoint (Fig.~\ref{fig:overview} bottom).
The triplane for $t=1$ is used to render the input frame, the frontal novel view image (used for supervision), and the groundtruth image for the sampled novel viewpoint.
The groundtruth novel view and the frontal novel view are only used for supervision.
In order to learn to fuse images under different lighting conditions, we also apply two separate color space augmentations to images at $t=0$ and $t=1$, each involving random alterations to brightness, contrast, saturation, and hue. 

\noindent\textbf{Shoulder Augmentation.}
It is important that 3D portraits include shoulders for improved realism and inclusion of body language in telepresence. 
In order for our model to learn to fuse images with different shoulder poses, we generate synthetic data of the same person with various shoulder rotations. 
However, Next3D does not provide control over shoulder rotation, and it is difficult to manipulate triplanes due to their implicit nature. 
Therefore, we propose to perform shoulder rotation augmentation during volume rendering.
Please refer to the supplementary for a visualization of this process.
In essence, we warp camera rays during volume rendering to simulate shoulder movement in the rendered image without having to modify the Next3D triplane. 
Through this augmentation, we synthesize various shoulder poses in rendered 2D images without modifying the Next3D triplane. 

\noindent\textbf{Pseudo-Groundtruth Triplanes.} \label{sec: pseudo_gt}
As a result of the shoulder augmentation, the 2D renderings involve changes that are not present in the original Next3D triplanes. This means that the Next3D triplanes cannot be used as direct supervisory signals. 
On the other hand, LP3D often generates reasonably accurate triplanes on frontal view images.
Therefore, to provide direct supervision signals to both the Triplane Undistorter $U$ and Fuser $F$ modules, we use a frozen LP3D to predict pseudo-groundtruth triplanes $T_{frontalGT}$ from the frontal novel view of $t=1$ (see bottom of Fig.~\ref{fig:overview}). 

\begin{figure*}
\vspace{-10pt}
  \centering
   \includegraphics[width=0.7\textwidth]{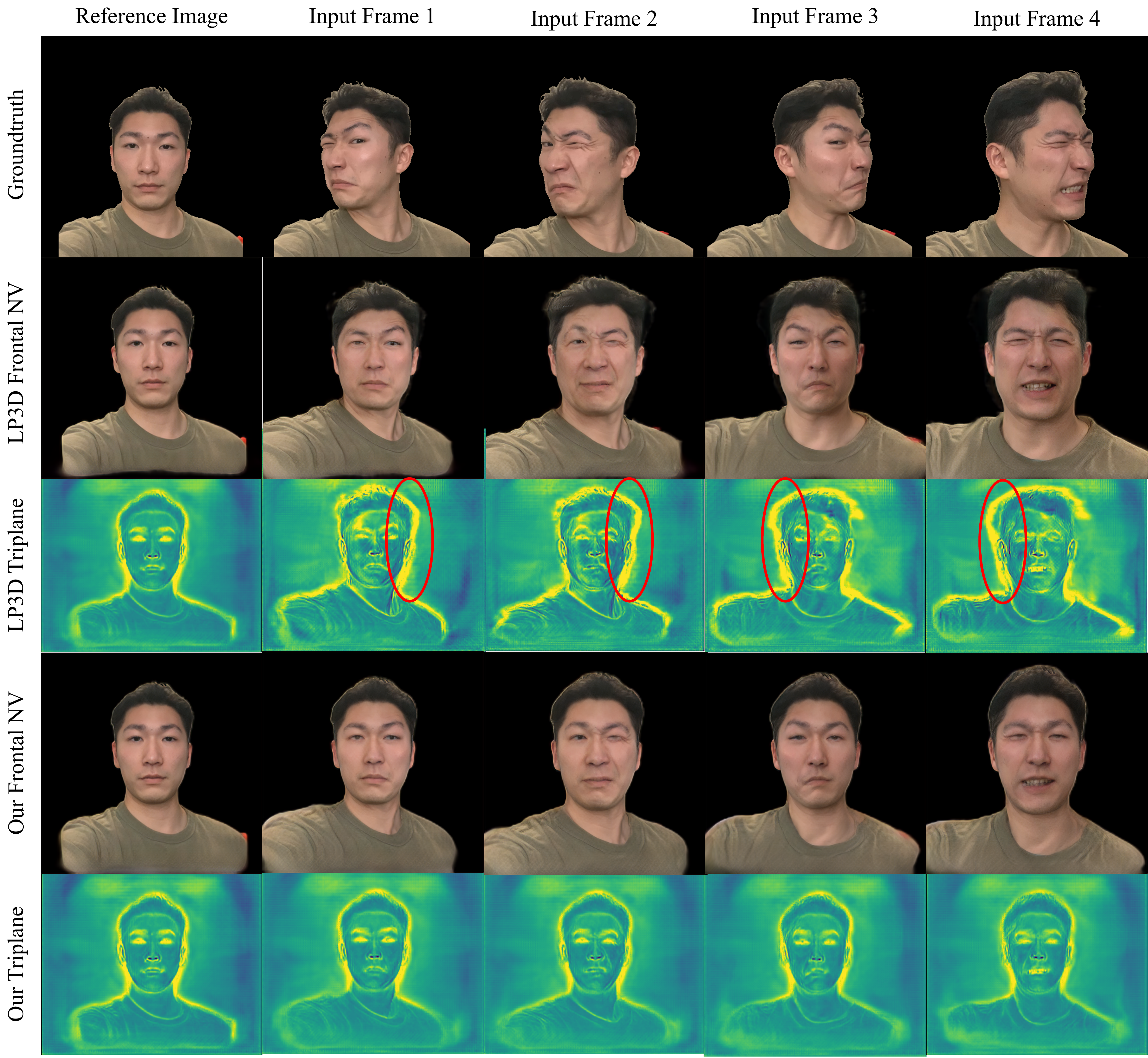}
   \vspace{-5pt}
   \caption{\textbf{View-Dependent Distortion:} \textit{Top}: inputs to our model and LP3D. \textit{Second \& Third Rows:} LP3D's reconstructions varies greatly under challenging viewpoints, showing predictable pattern of artifacts including abnormally strong activations on the side being captured (red circle), as well as geometric distortion along the view direction of the camera. 
   We refer to this phenomenon as "View-Dependent Distortion". \textit{Fourth \& Fifth Row:} Our method removes such artifacts and achieves better coherence.}
   \label{fig:distortion}
   \vspace{-15pt}
\end{figure*}

\subsection{Removing Distortion and Preserving Identity}
\label{sec:undistorter}
Our model uses a frozen pretrained LP3D to first predict a raw triplane $T_{raw}$ from an input video frame $I_{in}$.
Even though LP3D excels at faithfully reconstructing the 2D image from the input view, the quality of the actual 3D reconstruction 
is highly dependent on the person's head pose in the input image. 
For example, when the user is captured from the sides, LP3D often produces undesired artifacts such as incorrect identity, distortion along the camera's viewing direction (Fig.~\ref{fig:distortion}), and artifacts on the side of the camera (Fig.~\ref{fig:teaser} top). 

Visualizing the triplanes gives more insight to this problem and potential solutions. 
As shown in Fig.~\ref{fig:distortion}, when the person is captured from the sides, the raw triplanes $T_{raw}$ often exhibit abnormally strong activations on the side being captured, as well as geometric distortion along the view direction of the camera.
For example, for "Input Frame 1" and "Input Frame 2" columns, the camera captures the person from his left, and the LP3D triplanes show strong activation on the left side of the person (red circles), and the triplane and resulting renderings are also distorcted in the horizontal direction.
This phenomenon can be ascribed to the inherent ambiguity of single-image reconstruction. 
At the same time, we also notice that LP3D often work well with frontal views, which provide more complete identity information and less occlusion than side views.
Therefore, to reduce the single-image ambiguity, we use LP3D to reconstruct a personal triplane prior $T_{prior}$ from a frontal image, and we use $T_{prior}$ to constrain reconstructions of subsequent video frames.

A simple way to leverage the personal prior is to input both the triplane prior and current input frame into a neural network and rely on large scale training and data to help the model learn to generate coherent reconstruction.
However, we later show in Table.~\ref{table:ablation} (row "Only Fuser") that this approach turns out to be not very effective.
We find that it is easier for a network to correct the 3D reconstruction by warping the triplane rather than by generating a new triplane.
Therefore, we devise a Triplane Undistorter $U$ (see the pipeline Fig.~\ref{fig:overview}) that learns to correct the raw triplane $T_{raw}$ using triplane prior $T_{prior}$ as a reference and produces an undistorted triplane $T_{undist}$:
\begin{equation}
    \vspace{-1mm}
    U(T_{raw}, T_{prior})=T_{undist} \in \mathbb{R}^{3\times32\times256\times256}\ .
    \vspace{-1mm}
\end{equation}
More specifically, the Undistorter $U$ is based on the optical flow architecture from SPyNet\cite{ranjan2017optical}. 
SPyNet was originally developed to estimate dense optical flow to warp a source RGB image $I_{src}$ to a target image $I_{tgt}$ by iteratively estimating warping fields in a coarse-to-fine fashion.
We find that the same architecture is effective in predicting an undistortion flow map $T_{flow} \in \mathbb{R}^{3\times32\times256\times256}$, that reduces the distortion in $T_{raw}$:
\begin{gather}
    T_{flow} = SPyNet(T_{raw}, T_{prior}),\\
    T_{undist}=Warp(T_{raw}, T_{flow}).
    \vspace{-1mm}
\end{gather}

Notice that \textbf{the undistortion process is not optical flow prediction}. 
While a flow estimator would predict a flow that aligns the two inputs, Undistorter $U$ does not align the two inputs, \ie it does not warp $T_{raw}$ to $T_{prior}$ or vice versa. 
Instead, it merely uses $T_{prior}$ as the conditioning input to predict a correction warping to $T_{raw}$, producing $T_{undist}$. 
$T_{undist}$ is supervised by the pseudo-groundtruth triplane $T_{triplaneGT}$ (Sec.~\ref{sec:data}) via a triplane loss:
\begin{equation}
    \vspace{-1mm}
    L_{undist} = L_1(T_{undist}, T_{triplaneGT}).
\end{equation}

\subsection{Incorporating the Personal Triplane Prior via Triplane Fusion}
\label{sec:fuser}
As the user moves around in the video, different parts of their head become occluded. 
To recover occluded areas in the input frame and further stabilize the subject's identity across the video, our Fuser $F$ enhances the reconstruction by incorporating a personal triplane prior $T_{prior}$ lifted from a frontal reference image. 
Triplane priors $T_{prior}$ are essential to this process because the currently occluded areas are often visible in frontal images and $T_{prior}$, and frontal images also provides rather complete information about the person's identity and facial geometry, beneficial to stable identity reconstruction.
We feed both the undistorted triplane $T_{undist}$ and the triplane prior $T_{prior}$ to the Triplane Fuser $F$ to produce the final fused triplane $T_{fused}$ (see the pipeline Fig.~\ref{fig:overview}). 

During this process, it is important that the Fuser $F$ preserves the visible information in the input frame in order to accurately reconstruct dynamic conditions such as lighting changes.
Therefore, we explicitly predict a 3D visibility map for the input frame by estimating a visibility triplane $T^{raw}_{vis} \in \mathbb{R}^{3\times128\times128}$ for $T_{raw}$, \ie one visibility map for each plane, using a a 5-layer ConvNet.
$T_{raw}$ and $T^{raw}_{vis}$ are then concatenated and undistorted together before being input to the Fuser $F$ alongside $T_{prior}$ and its visibility triplane $T^{prior}_{vis}$. 
In this way, Fuser $F$ preserves visible facial regions in $T_{raw}$ and can recover the occluded regions using the triplane prior $T_{prior}$.

To train the visibility predictor, we calculate the visibility loss $L_{vis}$ as the $L_1$ distance between the predicted visibility triplanes ($T^{raw}_{vis}$ and $T^{prior}_{vis}$) and the groundtruth visibility triplanes ($T^{raw}_{visGT}$ and $T^{prior}_{visGT}$):
\begin{equation}
L_{vis} = L_1(T^{raw}_{vis}, T^{raw}_{visGT}) + L_1(T^{prior}_{vis}, T^{prior}_{visGT}).
\end{equation}

We compute the groundtruth visibility triplanes ($T^{raw}_{visGT}$ and $T^{prior}_{visGT}$) by first rendering a triplane into a depth image from the its input viewpoint. 
Then, we lift the depth map into a point cloud and project the points back onto the three planes in the triplane. 
The resulting visibility triplane $T_{visGT}$ is thus 1 for pixels where there is a point projection, and 0 otherwise. 
In this process we also calculate an occlusion mask $T^{raw}_{occMask} \in \mathbb{R}^{3\times 256\times 256}$ for the current frame $T_{raw}$ as the difference between the visibility $T^{raw,frontal}_{vis}$ of the frontal view and $T^{raw}_{vis}$ of the input view.
Please see the supplement for the full pipeline that includes the visibility calculation, visualization of the visibility triplanes, and training details. 

To supervised the Fuser $F$, we calculate the fusion loss $L_{fusion}$ as the $L_1$ loss between the fused triplane $T_{fused}$ and the pseudo-groundtruth triplane $T_{frontalGT}$. We also upweight the occluded regions via the occlusion mask $T^{raw}_{occMask}$:
\begin{equation}
L_{fusion} = Mean(\|T_{fused} - T_{triplaneGT}\| (1 + T_{visGT} + T_{occMask})) 
\end{equation}

We use the Recurrent Video Restoration Transformer (RVRT)~\cite{rvrt} as the backbone of our Fuser $F$ because of its memory efficiency. 
We replace the final summation skip connection in RVRT with convolutional skip connection because the summation skip connection prevents effective learning.
This is because the original RVRT was designed to correct local blurriness and noises in a corrupted RGB video, whereas our triplane videos exhibit structural distortion on a much larger scale and the summation skip connection thus limits the model's ability to correct the general structure.
We thus replace the summation with a small 5-layer ConvNet.

Lastly, note that both the Undistorter $U$ and the Fuser $F$ consist of 3 separate but identical fusers for each of the 3 planes because we find that using a single network to process all three planes causes collapse to 2D (please see supplementary for the visualization and discussions on such effects).

\subsection{Training Losses}
Our loss function is the summation of four loss terms that provide two types of supervision: (a) direct triplane space guidance used to supervise the undistortion process in the Undistorter $U$, the visibility prediction process, and the fusion process in the Fuser $F$; and (b) image space guidance for overall learning of high-quality image synthesis:
\vspace{-5pt}
\begin{equation}
L = w_{undist} L_{undist} + w_{vis} L_{vis} + w_{fusion} L_{fusion} + w_{render} L_{render} .
\end{equation}
$w_{undist}$, $w_{viz}$ , $w_{fusion}$, and $w_{render}$ are scalar weights for the different loss terms. \noindent $L_{render}$ is calculated as the perceptual loss $L_{LPIPS}$ between the groundtruth novel view $I_{GT}$ and the rendered novel view $I_{render}$:
\begin{equation}
L_{render} = L_{LPIPS}(I_{GT}, I_{render}) 
\end{equation}
\section{Results}
As discussed before, current methods like LP3D~\cite{trevithick2023} can overfit to the input viewpoints, but exhibits significant artifacts when synthesizing novel viewpoints for challenging input views like a profile picture.
Therefore, we need to evaluate the methods by examining their reconstruction across multiple viewpoints instead of only from the input view as done previously.

\subsection{Metrics} \label{sec:metrics}
We measure the accuracy of reconstructed identities by calculating the ArcFace\cite{deng2018arcface} cosine distance between the $I_{render}$ and $I_{GT}$:
\begin{equation}
    ID = 1 - ArcFace(I_{render})\cdot ArcFace(I_{GT}).
\end{equation}

To measure the accuracy of reconstructed expressions, we use the NVIDIA Maxine AR SDK~\cite{maxineSDK} to measure the $L_2$ distance between expression coefficients $e_{render}$ of the rendered image and $e_{GT}$ of the groundtruth:
\begin{equation}
    Expr = L_2(Maxine(I_{render}), Maxine(I_{GT})).
\end{equation}

\begin{figure*}[t!]
  \centering
    \includegraphics[width=\textwidth]{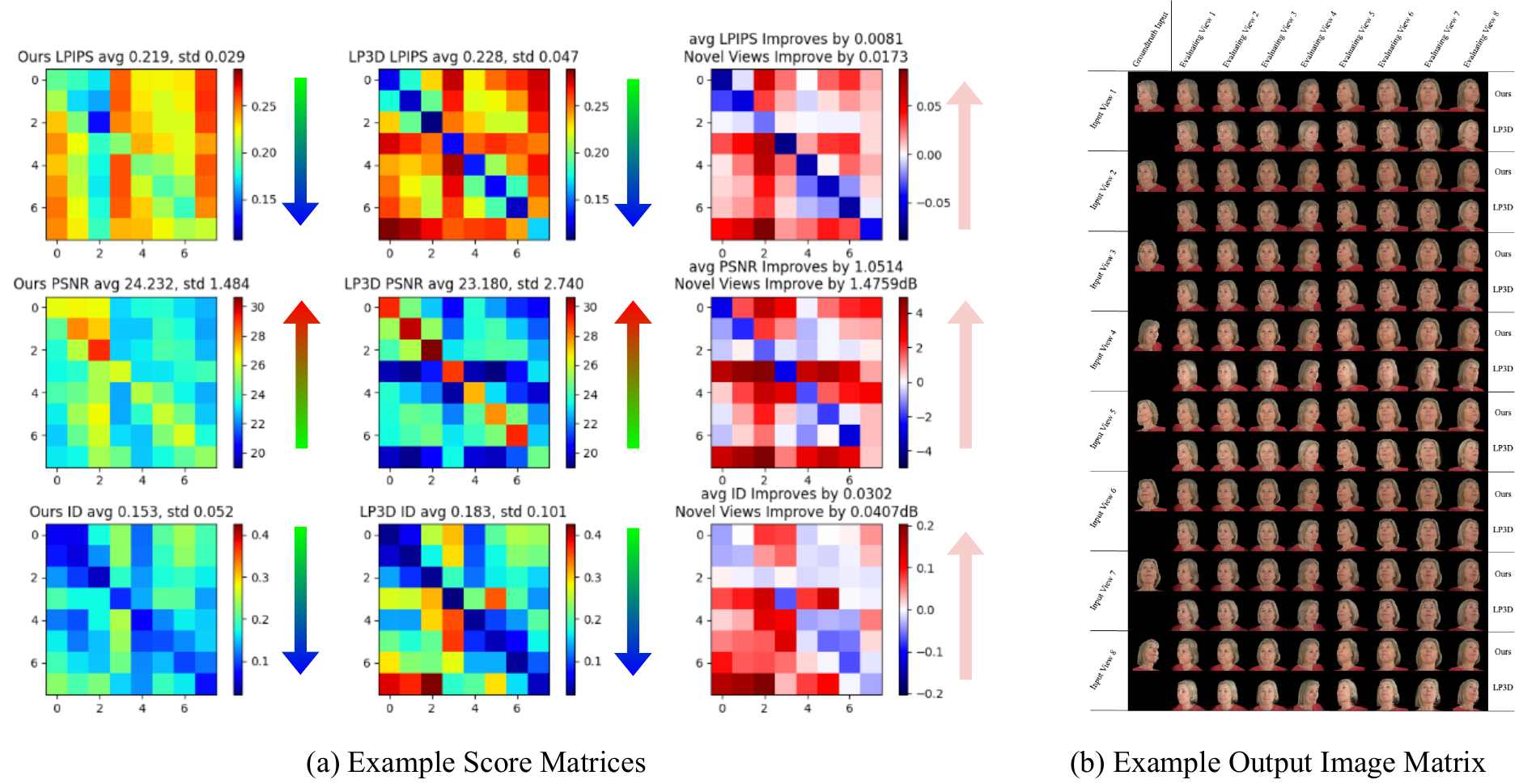}
   \caption{\textbf{Example Score Matrix and Output Image Matrix.} (a) We shown example Score Matrices S for the sequence "SEN-10-port$\_$strong$\_$smokey" in NeRSemble\cite{Nersemble}. Left 2 Columns: Ours and LP3D’s score matrices averaged over the test sequence. LPIPS (top) and ArcFace distance (bottom) are the lower the better, and PSNR (middle) is the higher the better. Right Column: red color represents improvement comparing to LP3D, and blue represents degradation. Our model achieves higher average performance and more uniform performance (lower standard deviation, more uniform color) whereas LP3D overfits to the input viewpoint and thus achieve higher performance for input views, but performs badly for novel views. (b) Example Output Image Images for LP3D and Ours on a frame in the sequence "EXP-3-cheeks+nose".
   }
   \label{fig:score_matrics}
\vspace{-20pt}
\end{figure*}

\noindent\textbf{Multi-View Evaluation of Single-View Reconstruction:} Due to the lack of 3D ground-truth for real-world data, prior methods are often evaluated on the input view reconstruction task using quantitative metrics like PSNR, 
whereas the novel view synthesis task often relies on visual assessments.
However, evaluating a reconstruction using only a single viewpoint can lead to ambiguities and inaccurate conclusions. 
For example, if the evaluation is only performed using the input viewpoint, then a method can overfit to the input view to achieve high numeric scores even if its reconstruction is highly inaccurate when rendered from novel viewpoints.
Moreover, single-view reconstruction methods can be heavily affected by the choice of input viewpoints. As shown in Fig.~\ref{fig:distortion}, different input views can lead to very different reconstructions.
Therefore, there are two variables crucial to the evaluation of single-view reconstruction methods: 
the choice of the input viewpoint and the choice of the evaluation viewpoint.
We thus propose new multi-view metrics that evaluate a model across different input-evaluation viewpoint combinations.
Using these new metrics, a method can only achieve high numeric performances when it consistently generates high-quality reconstructions regardless of the choice of input or evaluation viewpoints:

\noindent\textbf{Overall Synthesis Quality:} 
Given $N$ views in the dataset, we evaluate a method's average performance across different input-evaluation viewpoint combinations. More specifically, at each frame, each of the $N$ cameras is used as the input viewpoint to produce $N$ reconstructions in total, and each of the $N$ reconstruction is rendered and evaluated on the $N$ viewpoints, resulting in an $N\times N$ score matrix (Fig.~\ref{fig:score_matrics}). 
We use $N=8$ views in the NeRSemble\cite{Nersemble} dataset.
Thus, for a test sequence with $T$ frames, we generate a spatial-temporal score matrix $\textbf{S}^{T\times 8\times 8}$ for each of the metric (see the supplement for example visualization):
\vspace{-2mm}
\begin{gather}
    \textbf{S}_{t,i,j} = Metric(\textbf{I}_{render}^{t,i,j}, \textbf{I}_{GT}^{t,j}),1\leq i, j\leq N , 1\leq t\leq T.\\
    s = Mean(\{\textbf{S}_{t,i,j}\}).
    \vspace{-1mm}
\end{gather}
where $Metric(\cdot)$ can be LPIPS, PSNR, $ID$, and $Expr$ explained below. $\textbf{I}_{render}^{t,i,j}$ is the image rendered using camera $i$ as the input frame and camera $j$ as the output rendering view at frame $t$. $\textbf{I}_{GT}^{t,j}$ is the groundtruth frame captured by camera $j$ at frame $t$.
The Overall Synthesis Quality $s$ is thus the average over all score entries in $\textbf{S}$.
For a dataset of multiple test sequences, the final Overall Synthesis Quality is the average score of all sequences.

\noindent\textbf{Novel View Synthesis (NVS) Quality:} Novel View Synthesis Quality $s_{NV}$ is the average over all scores corresponding to novel view synthesis, \ie, the input view $i$ is different from output rendering view $j$:
\begin{equation}
s_{NV} = Mean(\{\textbf{S}_{t,i,j} | i\neq j, 1\leq t\leq T \}) .
\end{equation}

Additionally, it is also important to measure whether a method can authentically reconstruct dynamic real-life conditions in the video such as changes in lighting and shoulder poses. However, there is no existing multi-view in-the-wild portrait video dataset to support the evaluation of view synthesis quality. We thus qualitatively evaluate the methods on challenging in-the-wild portrait videos. Please see supplementary materials for image examples and video results.

\subsection{Dataset}
We quantitatively evaluate the methods on the NeRSemble~\cite{Nersemble} dataset, which is a high-quality multi-view portrait video dataset recorded with 16 calibrated time-synchronized cameras in a controlled studio environment. 
The images are captured at 7.1 MP resolution and 73 frames per second. 
There are 10 recordings in the test set, capturing a total of 10 individuals performing different expressions. 
NeRSemble provides us with the ability to evaluate the Overall Synthesis Quality and NVS Quality using the different input-evaluation viewpoint combinations. 
One of the 10 test sequences involves severe facial occlusion from hair that causes most of the methods' face trackers to fail for significant portions of the recording for many of the viewpoints. 
We thus leave out that sequence because the results would not be a reliable assessment of quality. 
We also use 8 roughly evenly separated cameras out of all 16 cameras during the evaluation. 

\subsection{Comparisons}
\begin{figure*}[t!]
  \centering
    \includegraphics[width=\textwidth]{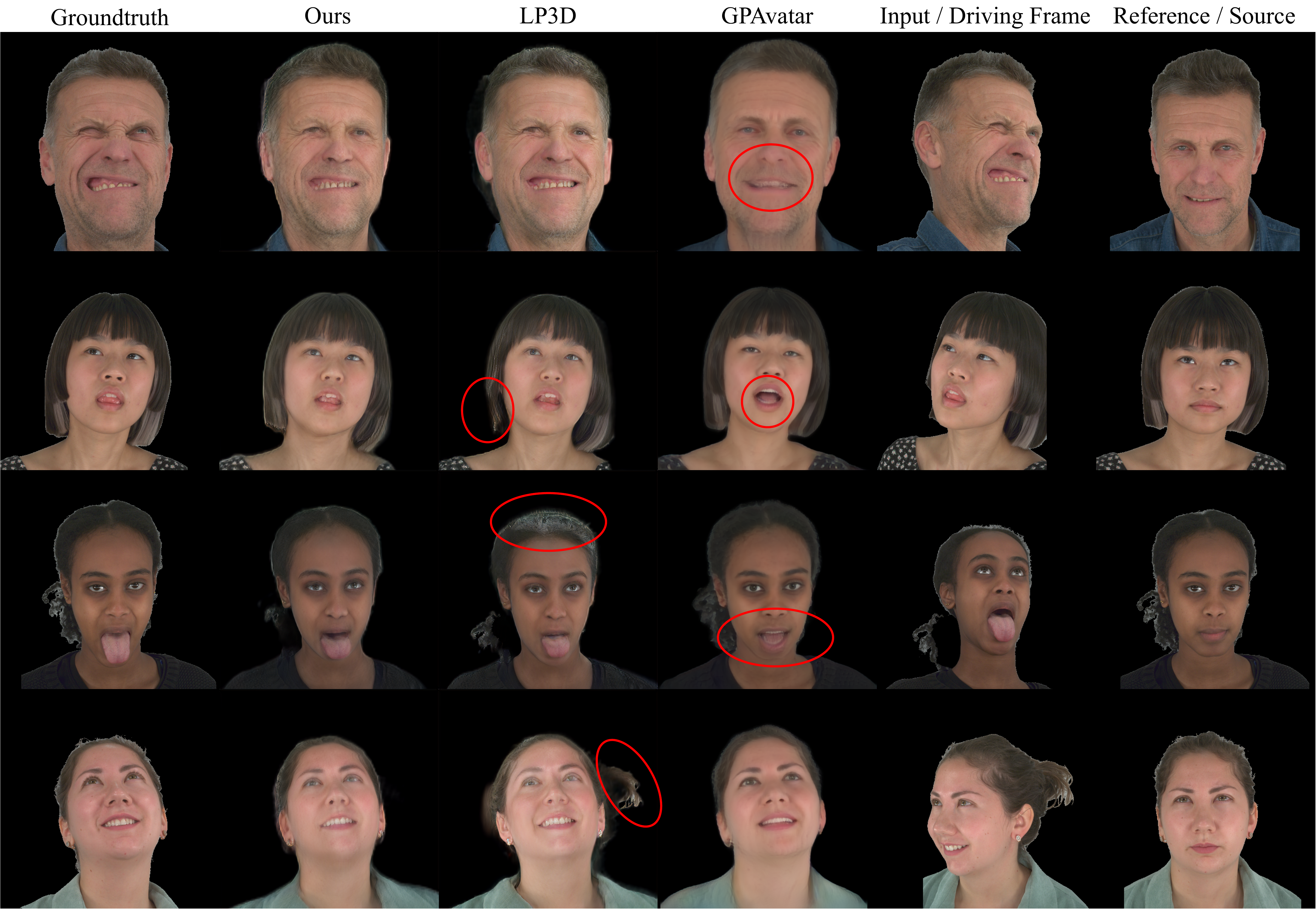}
   \caption{\textbf{Visual comparisons with baseline methods.} Our method strikes a balance between coherent reconstruction and faithful dynamic conditions like expressions. LP3D (third column) exhibits inconsistencies in identities, hairstyles, and artifacts (red circles). GPAvatar (fourth column) fails to capture challenging expressions (first row), new information not present in the reference image, (the stuck-out tongue in second and third rows), and identity of the person (last row).
   }
   \label{fig:comparison}
\end{figure*}

\noindent\textbf{Baselines} We compare our method with recent methods from 3 categories:

\textit{Reconstruction:} We evaluate LP3D\cite{trevithick2023} using the above protocol. We provide LP3D with the image from the input viewpoint and evaluate on all 8 viewpoints from the NeRSemble dataset. 

\textit{Reenactment:} Li \etal\cite{Li2023Oneshot} 
is able to reconstruct 3D portraits into a triplane from a reference image without test-time optimization, and they drive the reeconstruction via the frontal rendering of a 3DMM that modifies the expression in the original triplane. Concurrent to our work, GPAvatar\cite{chu2024gpavatar}  reconstructs 3D portraits by leveraging multiple source images and driving them through a FLAME\cite{FLAME:SiggraphAsia2017} mesh model. We test both methods in the self-reenactment setting. We use the first frame of the frontal camera in each NeRSemble test sequence as the reference image, and we drive it using videos of all 8 viewpoints. We evaluate GPAvatar using the same evaluation protocol as our method and LP3D. We evaluated Li \etal~\cite{Li2023Oneshot}'s approach using the input views as the only evaluating views, which are computed by the original authors, instead of all 8 views.

\textit{Inversion:} We also evaluate VIVE3D\cite{Fruehstueck2023VIVE3D}, which is a state-of-the-art 3D GAN inversion method for videos, and it can also perform semantic video editing. To perform inversion and evaluation on NeRSemble, VIVE3D's 3D GAN is first personalized using 3 frames from the input viewpoint video before inverting and rendering the reconstructed video from all 8 viewpoints.

Unfortunately, each of the above methods use different croppings of the face. We standardize the evaluation by re-cropping all methods to our cropping protocol, which is the largest of all. Please see the supplement for an additional table, where we evaluate the methods using different croppings around the face and arrive at conclusions consistent with Table.~\ref{table:split_column}.

\noindent\textbf{Quantitative Results} As mentioned before, we evaluate the methods using different input-evaluation viewpoint combinations, providing robust multi-view estimation for each of the metrics. Table.~\ref{table:split_column} shows that our model achieves state-of-the-art performance across all metrics versus recent works. 
Notably, LP3D is heavily affected by the input viewpoint, and our method is able to better preserve subject identity and expression (see Fig.~\ref{fig:distortion} and \ref{fig:comparison}). 
On the other hand, the reenactment methods struggle to capture authentic expressions because of the use of morphable face models, which have limited expressiveness. Moreover, they cannot faithfully reconstruct dynamic conditions (e.g. the stuck-out tongue in the second and third rows of Fig.~\ref{fig:comparison}) because they solely rely on information present in the source/reference images and do not incorporate new per-frame information. 
On the other hand, our method faithfully captures dynamic conditions and coherent reconstruction at the same time.

\begin{table}[h]
\centering
\begin{tabularx}{\textwidth}{l | c | c | c | Y Y | Y Y}
\hline
 \multirow{2}{*}{Method} & \multirow{2}{*}{Type} & \multirow{2}{*}{Expr$\downarrow$} & \multirow{2}{*}{ID$\downarrow$} & \multicolumn{2}{c|}{Overall Synthesis Quality} & \multicolumn{2}{c}{NVS Quality} \\
& & & & PSNR$\uparrow$ & LPIPS$\downarrow$ & PSNR$\uparrow$ & LPIPS$\downarrow$\\
\hline
Li \etal \cite{Li2023Oneshot} & reenact & 0.2657 & 0.2410 & 18.5733 & 0.2546 & 18.2020 & 0.2624 \\  
GPAvatar\cite{chu2024gpavatar} & reenact & 0.2041 & \underline{0.2074}  & 21.9487 & 0.2334 & \underline{21.9487} & \underline{0.2334} \\
VIVE3D\cite{Fruehstueck2023VIVE3D} & invert & 0.2900 & 0.3951 & 18.5771 & 0.2593 & 18.1449 & 0.2710 \\  
LP3D\cite{trevithick2023} & recon & \underline{0.1676} & 0.2154 & \underline{22.3309} & \underline{0.2232} & 21.5246  & 0.2374\\
Ours & recon & \textbf{0.1584} & \textbf{0.1865}  & \textbf{22.7695} & \textbf{0.2189} & \textbf{22.4395} & \textbf{0.2240}\\
\hline
\end{tabularx}
\caption{\textbf{Comparison on Nersemble~\cite{Nersemble}:} Our evaluation protocol (Sec.~\ref{sec:metrics}) utilizes multi-view groundtuth to evaluate each model. Under this robust evaluation, our method achieves state-of-the-art performance across all metrics. Our method achieves the best view synthesis accuracy and robustness to input viewpoints ("Overall Synthesis Quality" $\&$ "NVS Quality") while accurately capturing the identity and expression.}
\label{table:split_column}
\end{table}

\begin{table}[h]
\centering
\begin{tabularx}{\textwidth}{l | c | c |Y | Y | Y Y | Y Y}
\hline
 \multirow{2}{*}{Method} & \multirow{2}{*}{$U$} & \multirow{2}{*}{$F$} & \multirow{2}{*}{PSNR$\uparrow$} & \multirow{2}{*}{LPIPS$\downarrow$} & \multicolumn{2}{c|}{Input View Variation} & \multicolumn{2}{c}{Novel View Variation}\\
& & & & & PSNR$\downarrow$ & LPIPS$\downarrow$ & PSNR$\downarrow$ & LPIPS$\downarrow$ \\
\hline
LP3D\cite{trevithick2023} & 	\ding{55} & \ding{55} & \underline{22.3309} & 0.2232 & 1.0248 & 0.0152 & 2.2002 & 0.0532 \\
\hline
\multirow{3}{*}{Ours} & \checkmark & \ding{55} & 22.1963 & 0.2212 & 0.9069 & 0.0086 & 1.6990 & 0.0382 \\
 & \ding{55}& \checkmark & 22.2650 & \underline{0.2225} & \underline{0.5593} & \underline{0.0062} & \textbf{1.3150} & \textbf{0.0285} \\
 & \checkmark & \checkmark & \textbf{22.7695} & \textbf{0.2189} & \textbf{0.2453} & \textbf{0.0045} & \underline{1.3829} & \underline{0.0372} \\ 
\hline
\end{tabularx}
\caption{\textbf{Ablation studies.} We test two variations of our models (1) adding only the Undistorter $U$ to LP3D (row 2), and (2) adding only the Fuser $F$ (row 3). We show that simply adding each component does not lead to improvement. However, they complement each other and substantially improves the accuracy to the reconstruction (Novel View Variation) as well as the robustness to challenging input viewpoints (Input View Variation).}
\label{table:ablation}
\end{table}

\noindent\textbf{Ablations} We additionally evaluate two variations of our model: (a) with only the Triplane Undistorter added to LP3D (Table.~\ref{table:ablation}, row 2), and (b) with only the Triplane Fuser $F$ added to LP3D (row 3). 
In addition to the PSNR and LPIPS metrics, we develop two new metrics: 

\noindent(a) \textbf{Novel View Variation (NVV)} 
We evaluate how much a method's reconstruction quality varies across different evaluation views. 
We quantify this as the standard deviation of performance across the $N$ evaluating views using the same input view, \ie horizontal rows $1\leq i\leq N$ of the score matrix $S$ (Tab.~\ref{table:ablation} second column from the right): 
\begin{equation}
NVV = Mean(\{Stddev(\{\textbf{S}_{t,i,j} | i\neq j\})\} | 1\leq i\leq N, 1\leq t\leq T\}) .
\end{equation}
(b) \textbf{Input View Variation (IVV)} 
We measure how much a method's reconstruction quality varies when using input viewpoints (Sec.~\ref{sec:metrics} second column from the right).
We quantify this variation as the average standard deviation of performance on the same evaluation view using different input views, \ie vertical columns $1\leq j\leq N$ of the score matrix $S$ (Tab.~\ref{table:ablation} first column from the right).  
\begin{equation}
IVV = Mean(\{Stddev(\{\textbf{S}_{t,i,j} | i\neq j\})\} | 1\leq j\leq N, 1\leq t\leq T\}) .
\end{equation}
We observe that the inclusion of the Undistorter module consistently improves the "Novel View Variation" and "Input Robustness" metrics versus LP3D, indicating better robustness to different input viewpoints and more consistent rendering quality across views. 
However, when only the Undistorter is added (Tab.~\ref{table:ablation} second row) the PSNR is reduced. 
This is likely because this model does not leverage the reference image to improve the reconstruction of occluded areas. 
Additionally, by only undistorting the reconstruction, the Undistorter-only model loses the ability to achieve higher average score (but also higher standard deviations) by simply overfitting to the input view.
Similarly, the Fuser-only (Tab.~\ref{table:ablation} second row) achieves better robustness to different input viewpoints and more consistent rendering quality across views, but lower PSNR score.
A likely cause is that without the Undistorter, the Fuser needs to overcome the challenge of fusing highly misaligned triplanes, where the person look drastically different in the raw triplane $T_{raw}$ and triplane prior $T_prior$, possibly inducing more blurriness and alignment artifacts that lower the PSNR performane. 
Overall the best performance is achieved by including both the Undistorter and Fuser because the two modules complement each other. The Undistorter corrects the distortion in the raw triplane and thus reduces the challenges in fusing misaligned triplanes, and the Fuser recovers the occluded areas in the raw triplane.

\section{Discussion}
\label{sec:discussion}
\noindent\textbf{Conclusion.}
Recognizing the individual limitations of per-frame single-view reconstruction and 3D reenactment methods, we presented the first single-view 3D lifting method to reconstruct a 3D photorealistic avatar with faithful dynamic information as well as temporal consistency, which marries the best of both worlds. We believe our method paves the way forward for creating a high-quality telepresence system accessible to consumers. 

\noindent\textbf{Limitation and future work.}
With our method, fusing an extreme side view with a very different expression to the reference view may result in blurry reconstruction due to ambiguity in triplane alignment. 
We use a single reference image, but incorporating multiple ones with different expressions and head poses could lead to further improvements. 
While we focus on a modifying triplanes, tuning the feedforward network itself to integrate information across multiple temporal frames could lead to further improvements. 
Finally, due to the additional components, our current run-time performance is slower than real-time, which could be improved in future work.

%
%
\bibliographystyle{splncs04}
\bibliography{main}
\title{Supplementary Material \\ Coherent 3D Portrait Video Reconstruction via Triplane Fusion \vspace{-20pt}} 

\titlerunning{Coherent 3D Portrait Video Reconstruction via Triplane Fusion}


\author{Shengze Wang\href{https://mcmvmc.github.io/}{\inst{1}}\inst{,2} \and
Xueting Li\href{https://sunshineatnoon.github.io/}{\inst{2}} \and
Chao Liu\href{https://research.nvidia.com/person/chao-liu}{\inst{2}} \and
Matthew Chan\href{https://matthew-a-chan.github.io/}{\inst{2}} \and
Michael Stengel\href{https://research.nvidia.com/person/michael-stengel}{\inst{2}} \and
Josef Spjut\href{https://josef.spjut.me/}{\inst{2}} \and
Henry Fuchs\href{https://henryfuchs.web.unc.edu/}{\inst{1}} \and
Shalini De Mello\href{https://research.nvidia.com/person/shalini-de-mello}{\inst{2}} \and
Koki Nagano\href{https://luminohope.org/}{\inst{2}}}

\authorrunning{S.~Wang et al.}

\institute{University of North Carolina at Chapel Hill \and
NVIDIA
}

\maketitle

In this supplement, we show additional visual results on in-the-wild (Sec.~\ref{sec:inthewild}) and NeRSemble datasets (Sec.~\ref{sec:morenersemble}); provide an explanation of the shoulder pose augmentation process including synthetic multi-view data generation using Next3D~\cite{sun2023next3d} (Sec.~\ref{sec:shoulder}); explain how visibility and occlusion calculations are performed in our method  (Sec.~\ref{sec:visibility}); visualize the outputs and score matrices that we use to calculate performance metrics (Sec.~\ref{sec:scoremat}); describe the cropping and training modifications made to the original LP3D (Sec.~\ref{sec:crop}); present three additional sets of quantitative results using different crops of the face (Sec.~\ref{sec:facecropresults}) and, lastly, show how jointly fusing the three planes can cause collapse to 2D (Sec.~\ref{sec:collapse}). 
We strongly recommend that readers \textbf{view the accompanying video} with this document for better assessment of the quality of the results of the various methods.

\vspace{-10pt}
\section{Additional Comparisons}
\label{sec:moreresults}
In this section, we show more qualitative comparisons between LP3D\cite{trevithick2023}, GPAvatar\cite{chu2024gpavatar}, VIVE3D\cite{Fruehstueck2023VIVE3D}, One-Shot-Avatar\cite{Li2023Oneshot} and our method. 
in Figs.~\ref{fig:inthewild1},~\ref{fig:inthewild2},~\ref{fig:inthewild3},~\ref{fig:comparison1}, and~\ref{fig:comparison3}. We highly encourage readers to view the \textbf{supplementary video}, which provides more visual comparisons. 
\subsection{In-The-Wild-Data.} \label{sec:inthewild}
\begin{figure}[!b]
\vspace{-34pt}
  \centering
    \includegraphics[width=0.75\textwidth]{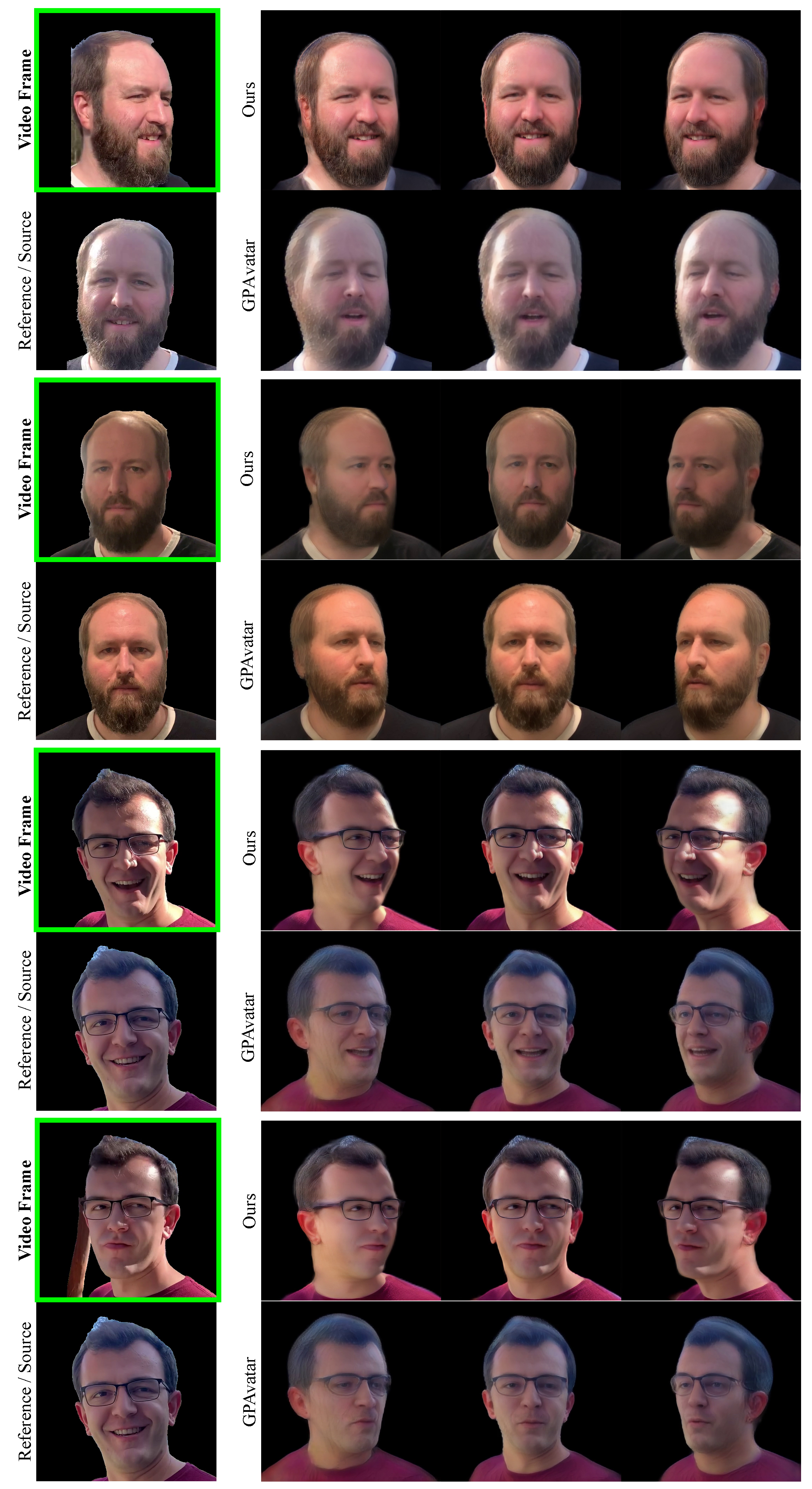}
   \caption{\textbf{In-the-wild Lighting (GPAvatar Vs. Ours):} Our method captures the dynamic lighting changes in the input video whereas GPAvatar fails to do so. Note that the output of the models should match the lighting and expression of input \textit{Video Frame} (\textcolor{ForestGreen}{GREEN} box).
   }
   \label{fig:inthewild1}
   \vspace{-10pt}
\end{figure}

\begin{figure}[!b]
\vspace{-34pt}
  \centering
    \includegraphics[width=0.76\textwidth]{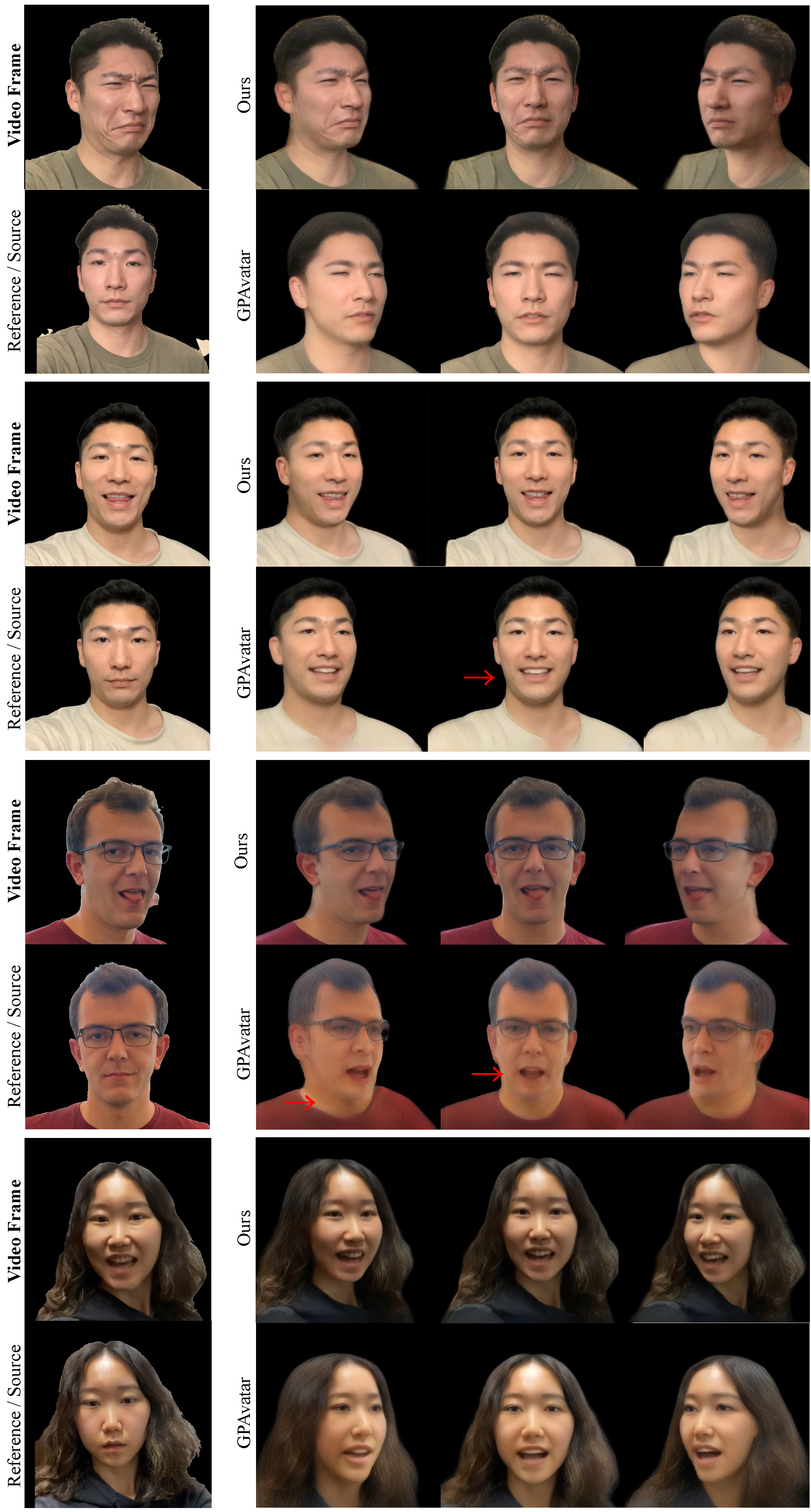}
   \caption{\textbf{In-the-wild Expression (GPAvatar Vs. Ours):} Our method more accurately captures human expressions in the input video whereas GPAvatar fail to reconstruct authentic expressions. Note that the output of the models should match the lighting and expression of input \textit{Video Frame}.
   }
   \label{fig:inthewild2}
   \vspace{-10pt}
\end{figure}

\begin{figure}[!b]
\vspace{-34pt}
  \centering
    \includegraphics[width=0.78\textwidth]{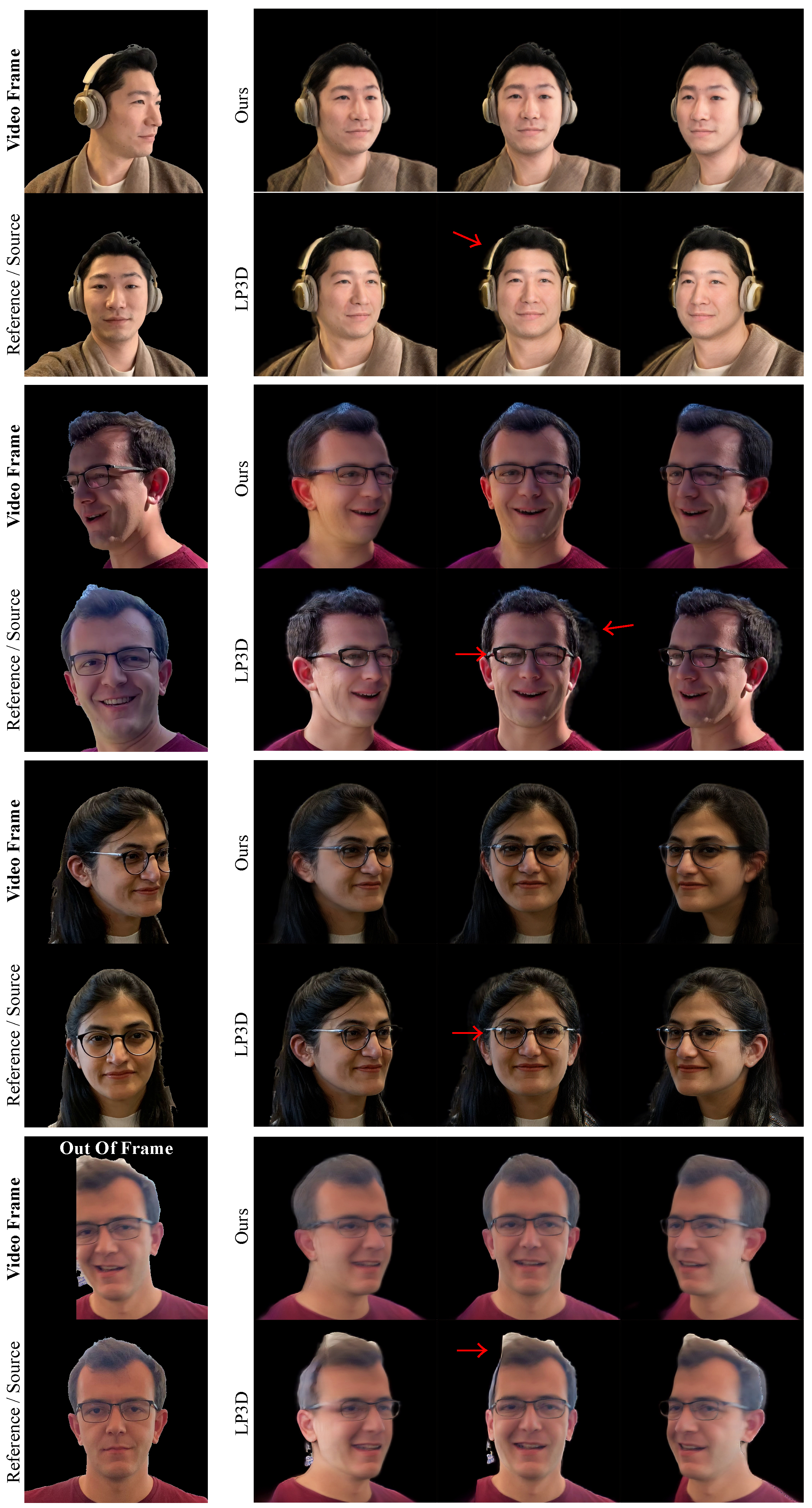}
   \caption{\textbf{In-the-wild Viewpoints (LP3D Vs. Ours):} Our method is more robust to variations in the input viewpoint whereas LP3D often performs poorly on rendering novels views that are far from the input view point. Note that the output of the models should match the lighting and expression of input \textit{Video Frame}.
   }
   \label{fig:inthewild3}
   \vspace{-10pt}
\end{figure}

In Figs.~\ref{fig:inthewild1},~\ref{fig:inthewild2}, and~\ref{fig:inthewild3}, we show results of GPAvatar, LP3D and our model on challenging in-the-wild test sequences. 
Since NeRSemble is a high-quality dataset captured in a controlled studio environment, it is different from real-life usage and limited in terms of lighting conditions, camera viewpoints, and motion blur.
Therefore, we capture people of different gender and race in daily environments like offices, apartments, and in outdoor open areas to evaluate the performance of different models in challenging in-the-wild situations.
The dataset includes 9 video sequences and 1 image set captured from iPhones, all of which are shown in this supplementary.
Our model is able to capture lighting changes (Figs.~\ref{fig:inthewild1}), maintain stable identity (Figs.~\ref{fig:inthewild3}), and remembering the user when their face is partially
out of the frame (Figs.~\ref{fig:inthewild3}, second row from the bottom), whereas LP3D shows temporal inconsistency (Figs.~\ref{fig:inthewild3}, red arrows); GPAvatar is not only unable to capture the live lighting condition of the user, but also fails to reconstruct their expressions accurately (Figs.~\ref{fig:inthewild1} and Figs.~\ref{fig:inthewild2}).

\newpage

\newpage

\newpage

\clearpage
\newpage
\subsection{Additional Results on NeRSemble.}\label{sec:morenersemble}
\begin{figure}[!b]
\vspace{-20pt}
  \centering
    \includegraphics[width=\textwidth]{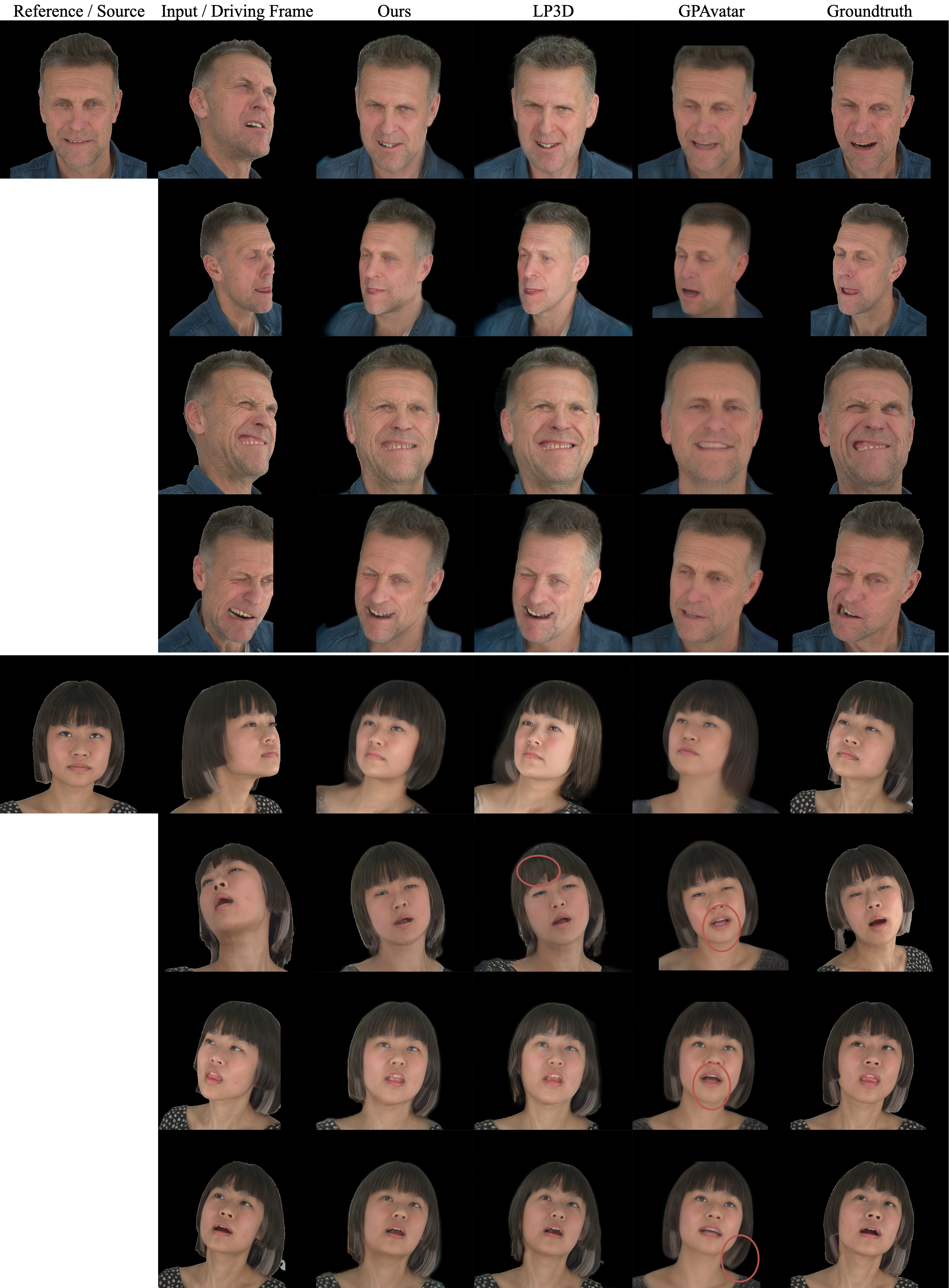}
   \caption{\textbf{Example comparisons on NeRSemble sequences}. Our model is able to capture extreme expressions and dynamics in hair movement (last row) while maintaining consistent identity despite viewpoint changes. 
   On the other hand, LP3D shows inconsistent identities and
   GPAvatar exhibits inaccurate expressions and significantly more blurry results. 
   GPAvatar also fails to reconstruct novel content such as the tongue (second last row) and different hair movement (last row). 
   The quality of expression reconstruction is best viewed in the video. 
   }
   \label{fig:comparison1}
\vspace{-10pt}
\end{figure}

We notice that, despite good numerical performance in terms of LPIPS and PSNR, a closer visual inspection of GPAvatar's results reveals that it is visually not as convincing as the two metrics indicate. This is because it renders dampened expressions (Fig.~\ref{fig:comparison1} top examples) and hallucinates parts of the face not present in the reference image (the inner mouth and tongue in Fig.~\ref{fig:comparison1} bottom example third row). 
LP3D is able to reconstruct nuanced facial expressions but struggles to maintain coherent identity when different viewpoints are used as inputs (see Fig.~\ref{fig:comparison1} top example first row).
Our model achieves both of these properties.

\subsection{VIVE3D \& Li \etal\cite{Li2023Oneshot}}
\begin{figure}[!t]
  \centering
    \includegraphics[width=\textwidth]{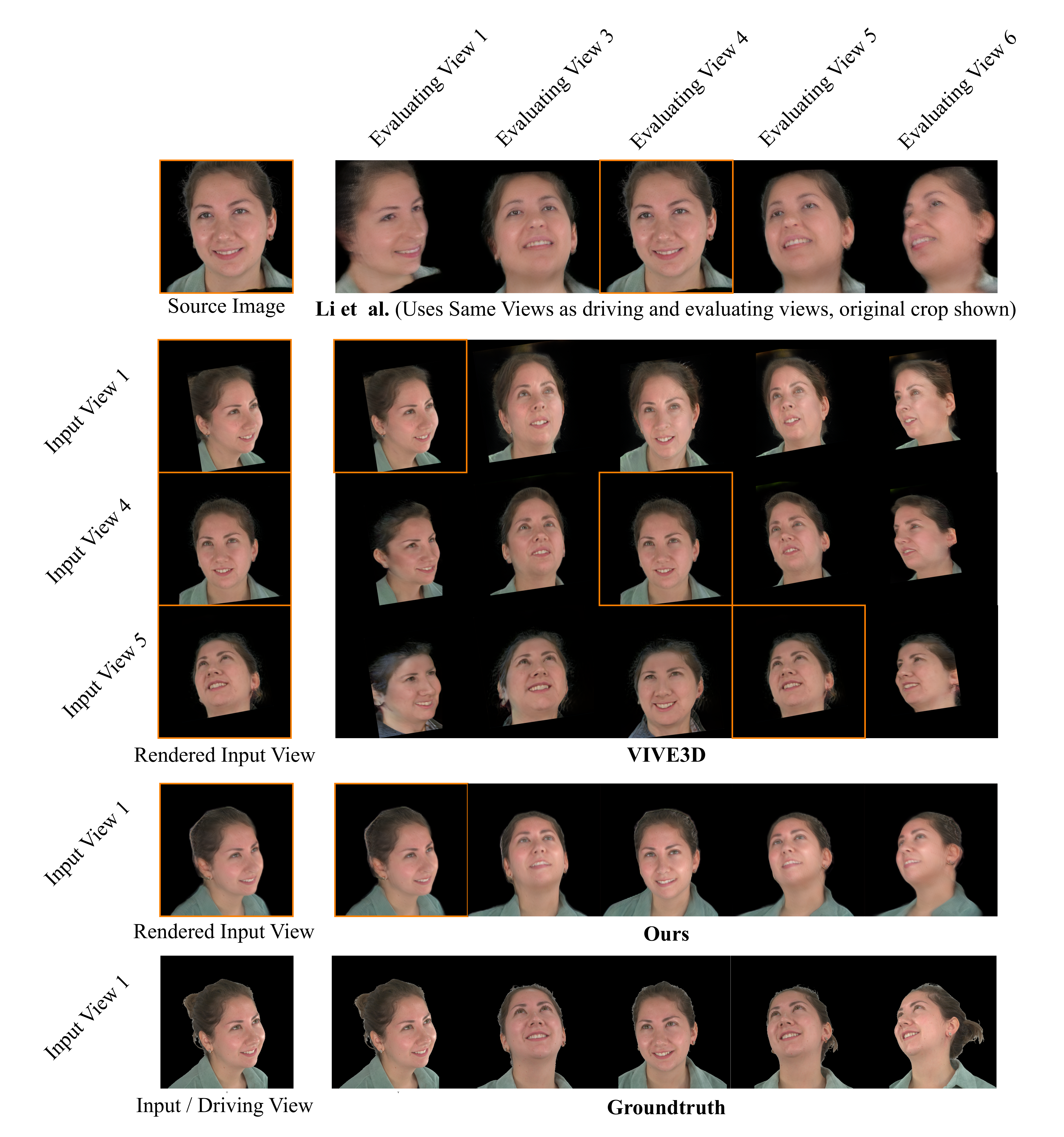}
    \vspace{-10pt}
   \caption{\textbf{VIVE3D~\cite{Fruehstueck2023VIVE3D} and Li \etal~\cite{Li2023Oneshot}:} \textit{"Li \etal\cite{Li2023Oneshot}" Row}: The authors kindly evaluated their methods for us. They evaluated the method on the same driving/input viewpoint (highlighted in orange) instead of all 8 viewpoints. 
   This method excels at frontal views but shows significant blurriness from the sides as well as unnatural expressions.
   \textit{"VIVE3D\cite{Fruehstueck2023VIVE3D}" Row}: VIVE3D is heavily affected by the input viewpoint. 
   It excels at reconstructing the input views but fails to reconstruct other viewpoints well.
   \textit{"Ours" Row}: Our method is able to achieve better reconstructions using the same input view as the other methods.
   We omit detailed results from VIVE3D and Li \etal in the main paper due to their less competitive results. 
   Images shown are at the original resolution.
   }
   \label{fig:comparison3}
\vspace{-10pt}
\end{figure}

In our main paper and supplement, we mostly omitted results from Li \etal~\cite{Li2023Oneshot} and VIVE3D~\cite{Fruehstueck2023VIVE3D} because of their less competitive results. 
As mentioned in the main paper, the authors of Li \etal~\cite{Li2023Oneshot} kindly performed evaluations for us. Different from other methods, the results are evaluated only on the input viewpoints instead of all 8 viewpoints for NeRSemble. 
In Fig.~\ref{fig:comparison3}, we show that this method excels at frontal views but shows significant blurriness from the sides as well as unnatural expressions.
On the other hand, VIVE3D is heavily affected by the input viewpoint. 
It excels at reconstructing the input views but fails to reconstruct other viewpoints well.
Comparing to the two methods, we achieve much better reconstruction.

\clearpage
\newpage
\newpage
\section{Shoulder Pose Augmentation}\label{sec:shoulder}
\begin{figure}[h]
  \centering
    \includegraphics[width=\textwidth]{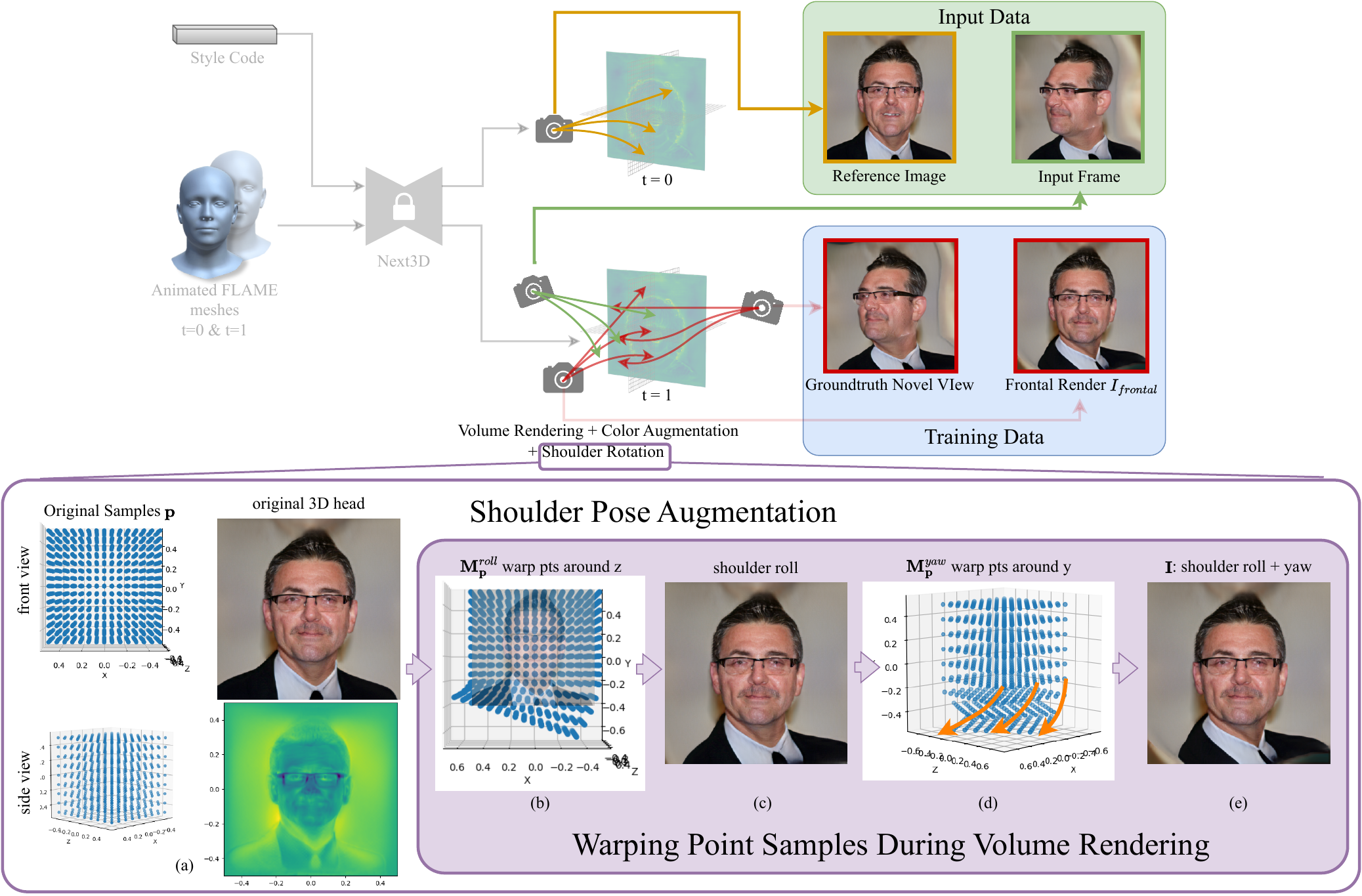}
   \caption{\textbf{Shoulder Augmentation.} Our data generator (Next3D\cite{sun2023next3d}) does not allow for control over shoulder poses. To enable our model to learn to fuse triplanes with different shoulder poses, we perform shoulder pose augmentation during volume rendering.
   }
   \label{fig:shoulder}
\vspace{-20pt}
\end{figure}

As shown in Fig.~\ref{fig:shoulder}, for training we generate 2 input images (\ie, a Reference Image and an Input Frame in the green box), and 2 groundtruth images using Next3D (in the blue box).
We use these images to train our triplane fusion module such that it learns to enhance the reconstruction of the input frame by leveraging a frontal reference frame. 
When used in practice, the input frame often contains shoulder rotations that are different from that of the reference frame.
It is important to reconstruct the varying shoulder pose in the input video because it conveys nuanced body language that is crucial to the perceived realism of an applications such as telepresence.

We utilize a pretrained 3D GAN, Next3D\cite{sun2023next3d}, as our training data generator.
However, Next3D does not allow us to synthesize different shoulder poses for the same person.
Since it is difficult to change the 3D geometry encoded in triplanes, we synthesize different shoulder poses in the 2D renderings by bending camera rays during volume rendering, \ie, by applying a warping field $\textbf{M}$ to the 3D points sampled. More formally, we apply the warp fields $\textbf{M}_\textbf{p}^{roll}$ and $\textbf{M}_\textbf{p}^{yaw}$ sequentially in order to transform the set of point samples $\textbf{p}$ used during volume rendering $R(\cdot)$. 
The final rendered image, $I$, thus uses the warped point $\textbf{p}'$ to sample the triplane $T$ during volume rendering $R(\cdot)$:
\begin{gather}
    \textbf{p}' = \textbf{M}_\textbf{p}^{yaw} \textbf{M}_\textbf{p}^{roll} \textbf{p} ,\\ 
    \textbf{I} = R(\textbf{p}', T).
    \label{eqn:shoulder}
\end{gather}
We show an overview of this shoulder augmentation process at the bottom of Fig.~\ref{fig:shoulder}.

In Fig.~\ref{fig:shoulder}(a), we show the original 3D head,
the Next3D triplane ($y$-axis upwards, $x$-axis to the right, $z$ out-of-the-plane), which ranges from -0.5 to 0.5 on all axes,
as well as uniform point samples that represent the 3D space before being warped. Then, we warp the camera point samples to achieve shoulder roll (Fig.~\ref{fig:shoulder}(b)).
The warping transform is only applied to the neck and shoulder regions, which are highly consistent in terms of position across Next3D triplanes. 
This is because 3D GANs like Next3D and EG3D\cite{eg3d2022} learn a canonical head space from 2D face crops of consistent sizes.
Therefore, we find that the neck and shoulder regions can simply be expressed by all point samples $\textbf{p}_{shoulder}=(x,y,z)$, where $y<y_{chin}$, where $y_{chin}=0.2$.

We rotate $\textbf{p}_{shoulder}$ around the top of the neck vertebrae, for which we heuristically use the origin as the rotation pivot.
Since a uniform rigid rotation would result in discontinuities, we apply increasingly larger rotations to the points based on their $y$ (vertical) coordinates.
Therefore, given a roll rotation angle $\theta_{base}$ for the base of the shoulder at $y_{base}=-0.5$, the roll rotation matrix $M_\textbf{p}$ for point $\textbf{p}$ can be calculated as:
\begin{gather}
    d_{chin} = \|y-y_{chin}\|, \\
    \theta_\textbf{p} = d_{chin} / \|y_{base}-y_{chin}\| \times \theta_{base},\\
    \textbf{M}_\textbf{p}^{roll} = 
    \begin{pmatrix}
        \cos(\theta_\textbf{p}) & -\sin(\theta_\textbf{p}) & 0 \\
        \sin(\theta_\textbf{p}) & \cos(\theta_\textbf{p}) & 0 \\
        0 & 0 & 1
    \end{pmatrix}.
\end{gather}

Similarly, given the yaw rotation angle $\phi_{base}$ for the base of the shoulder, the yaw rotation angle $\phi_\textbf{p}$ and matrix $\textbf{M}_\textbf{p}^{yaw}$ for point $\textbf{p}$ can be calculated as 
\begin{gather}
    \phi_\textbf{p} = d_{chin} / \|y_{base}-y_{chin}\| \times \phi_{base}, \\
    \textbf{M}_\textbf{p}^{yaw} = 
    \begin{pmatrix}
        \cos(\phi_\textbf{p}) & 0 & -\sin(\phi_\textbf{p}) \\
        0 & 1 & 0 \\
        -\sin(\phi_\textbf{p}) & 0 & \cos(\phi_\textbf{p}) \\
    \end{pmatrix}.
\end{gather}

The final rendered image, $I$, is thus generated by the volume rendering function $R(\cdot)$ with warped point samples $\textbf{p}'$ to sample the triplane $T$ using Eqns. (1) and (2).

\section{Visibility Estimation and Occlusion Masks}\label{sec:visibility}
\begin{figure}[h]
  \centering
    \includegraphics[width=\textwidth]{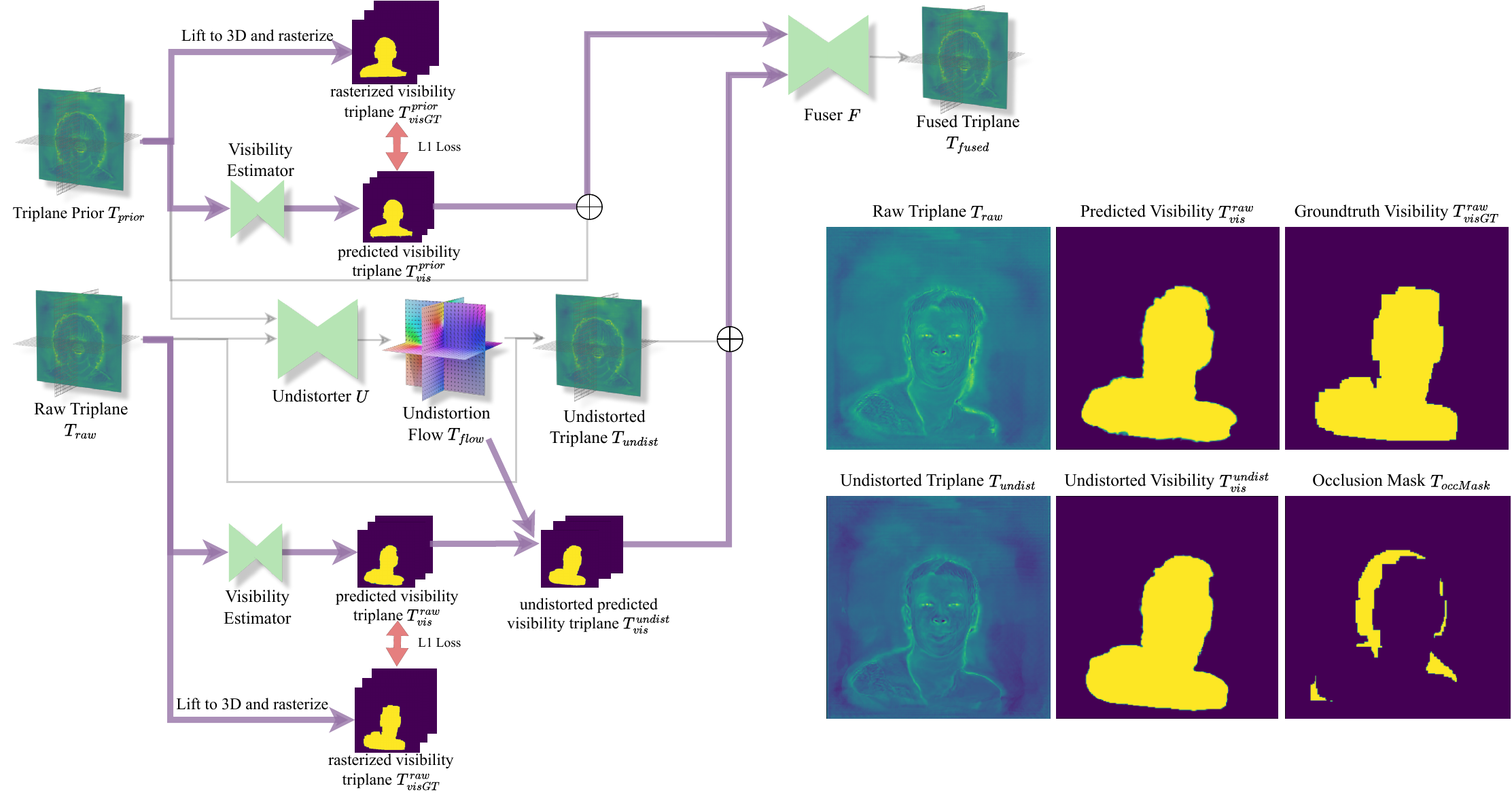}
   \caption{\textbf{Visibility Estimation:} We show the flow of visibility information in purple.}
   \label{fig:vis}
\vspace{-20pt}
\end{figure}

LP3D generates a complete triplane (and thus 3D portrait) from a single image, which inevitably contains occlusion.
For example, when the camera captures the person from the right, the right side of the face is visible and thus more reliable in the reconstruction whereas the left side of the face is occluded and thus is often inaccurately hallucinated by LP3D.
Therefore, to fuse reliable information from the input frame (\ie raw triplane $T_{raw}$) and the reference image (\ie triplane prior $T_{prior}$), it is important to inform the fuser $F$ about visible (and thus reliable) regions on the two triplanes.

In Fig.~\ref{fig:vis}, we show how we predict and leverage visibility information by high lighting the data flow of visibility information through our network in purple. 
First, our model estimates a predicted visibility triplane $T_{vis}^{raw}$ for the raw triplane $T_{raw}$.
Second, the visibility triplane is undistorted alongside $T_{undist}$ using $T_{flow}$.
Finally, the undistorted visibility triplane $T_{vis}^{undist}$ informs the Fuser $F$ about the visibility/reliability of different regions in $T_{undist}$ and allows for better fusion.

\noindent\textbf{Visibility Mask Triplane.} 
There are various ways to compute the visibility information for a triplane. 
For simplicity, we approximate the actual visibility masks through a rasterization approach: Given a triplane $T$ and its input camera pose $C$, we generate a pseudo-groundtruth visibility triplane $T_{visGT}$ by first rendering triplane $T$ into a depth map via volume rendering from camera $C$. 
We then lift the depth map into a 3D point clouds and rasterize the point cloud back onto the triplane by orthographically projecting the points onto the xy, yx, and xz-planes. 
The final visibility mask is 1 where points are rasterized and 0 where none is rasterized.
Therefore, for triplane prior $T_{prior}$ and the raw triplane $T_{raw}$, we can calculate pseudo-groundtruth visibility triplanes $T_{visGT}^{prior}$ as well as $T_{visGT}^{raw}$.

However, this process is expansive due to the volumetric rendering used for depth map generation, we thus develop a Visibility Estimator to directly predict the visibility triplanes. 
Our Visibility Estimator is a 5-layer ConvNet that predicts visibility maps $T_{vis}^{prior},T_{vis}^{raw} \in \mathbb{R}^{3\times 1\times 256\times 256}$ 
from the triplane prior $T_{prior}$ and raw triplane $T_{raw}$, respectively.

The two visibility maps are concatenated with $T_{prior}$ and $T_{undist}$ before being input into the Triplane Fuser $F$. 
In Fig.~\ref{fig:vis}(right), we show an example of the raw triplane $T_{raw}$ and its predicted visibility triplane $T_{vis}^{raw}$, undistorted triplane $T_{undist}$, its visibility triplane $T_{vis}^{undist}$, and the pseudo-groundtruth visibility triplane $T_{visGT}^{undist}$. 

\noindent\textbf{Occlusion Mask Triplane.} 
In addition to providing the Fuser $F$ with helpful information about visibility, it is also beneficial to emphasize the reconstruction of occluded areas during training because it encourages the model to leverage the frontal reference image for the reconstruction of occluded areas. 
To achieve this we use an occlusion mask triplane $T_{occMask} \in \mathbb{R}^{3\times 1\times 256 \times 256}$ to upweight the triplane loss on occluded areas on the triplane indicated by the mask (see main paper Sec.~ 4.4). 
$T_{occMask}$ is calculated as
the difference between the visibility triplane $T_{visGT}^{raw}$ of the raw triplane versus the much more complete visibility triplane $T_{visGT}^{prior}$ of the triplane prior.

The Visibility Estimator is supervised via an $L_1$ loss between the predicted visibility triplane and its groundtruth as:
\begin{gather}
    L_{vis} = L_1(T_{vis}^{raw}, T_{visGT}^{raw}) + L_1(T_{vis}^{prior}, T_{visGT}^{prior}).
\end{gather}

\begin{figure}[t!]
  \centering
    \includegraphics[width=\textwidth]{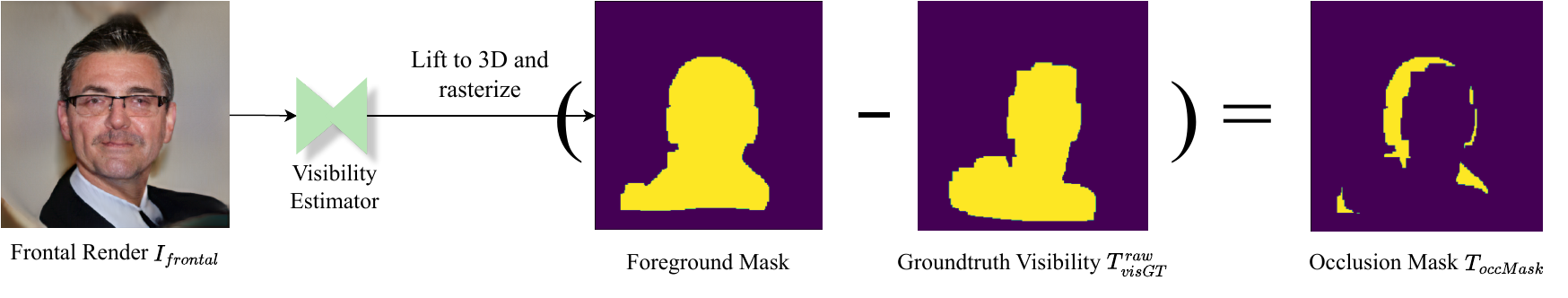}
   \caption{\textbf{Occlusion Calculation:} The occlusion map is the difference between the approximated foreground mask and the groundtruth visibility triplane $T_{vis}^{raw}$}.
   \label{fig:occ}
\end{figure}

\section{Visualization of Score Matrix}\label{sec:scoremat}
In Fig.~\ref{fig:scoremat}, we show example Score Matrices $\textbf{S}$ for the NeRSemble dataset's sequence "SEN-10-port\_strong\_smokey". 
Each cell $\textbf{S}_{i,j}$ represents the score of the reconstruction using view $i$ as the input and view $j$ as the evaluation view.
Our model achieves higher average as well as more uniform performance, because it has a lower standard deviation and hence more uniform color. 
Additionally, our model achieves improvements for a majority of the cells (input-evaluation view combinations).

\clearpage
\newpage
\begin{figure}[h]
  \centering
    \includegraphics[width=0.8\textwidth]{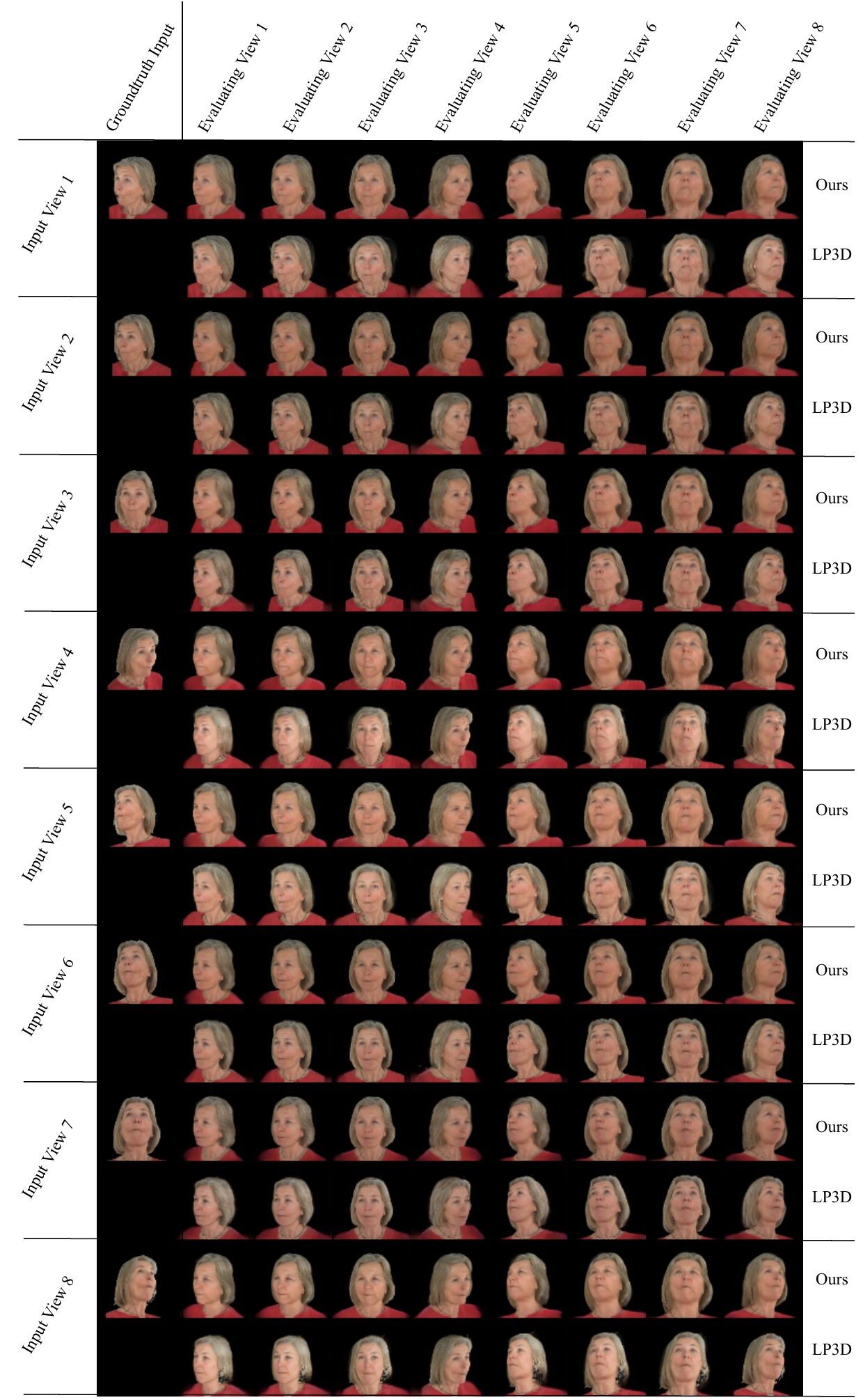}
   \caption{\textbf{Output Matrices of LP3D and Our Method:} 
   We show an example output matrix from a frame in a NeRSemble test sequence. 
   Each row represents the process of creating a 3D head from the input view (left), and evaluating the reconstruction by rendering on all 8 viewpoints.
   The images in this $2\times 8\times 8$ output matrix are $512\times512$ each, leading to a large image. The shown output matrix is downsampled for visualization.}
   \label{fig:scoremat}
\end{figure}

\clearpage
\newpage
\begin{figure}[h]
  \centering
    \includegraphics[width=\textwidth]{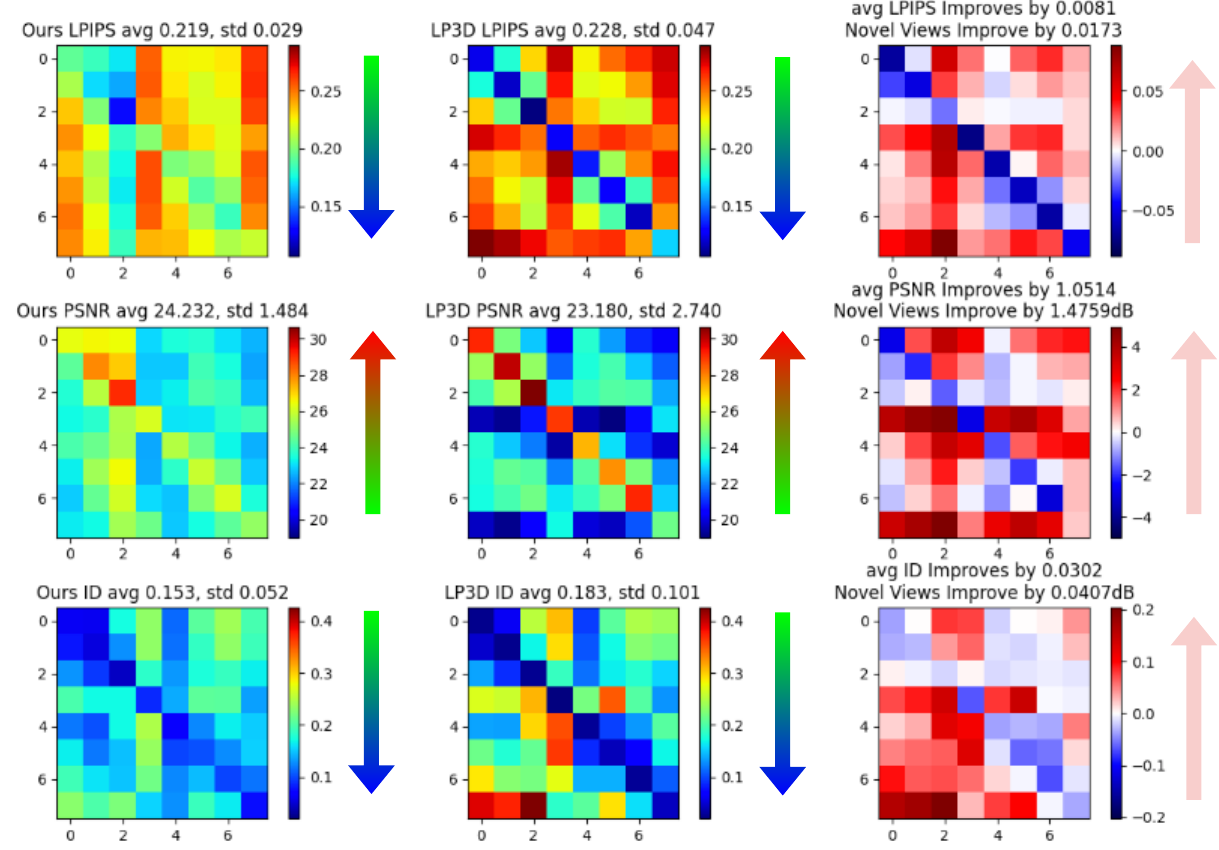}
   \caption{\textbf{Score Matrix:}
   We show example Score Matrices $\textbf{S}$ for the sequence named "SEN-10-port$\_$strong$\_$smokey". \textit{Left 2 Columns}: Ours' and LP3D's score matrices averaged over the test sequence. LPIPS (top row) and ArcFace ID cosine distance (bottom row) are the lower (greener/bluer) the better, and PSNR (bottom row) is the higher the better (greener/redder)). 
   \textit{Right Column}: red color represents improvement compared to LP3D, and blue represents degradation. 
   Notice that changes in LPIPS and ArcFace ID losses are negated such that positive numbers (red) reflect positive changes.
   Our model achieves higher average performance as well as more uniform performance (lower standard deviation, more uniform color) whereas LP3D overfits to the input viewpoint and thus achieve higher performance for input views, but performs badly for novel views.
   }
   \label{fig:scoremat}
\end{figure}
\section{Cropping Modifications to LP3D}\label{sec:crop}
Our implementation of LP3D mostly follows the original LP3D with a few small modifications. 
The original LP3D was trained for tight cropping around faces for FFHQ corresponding to the normalized focal length of 4.26 in EG3D~\cite{eg3d2022}. To capture the whole head including shoulders, we increased the field of view and retrained the LP3D with the normalized focal length of 3.12. 

\section{Performance on Face-only Crops}\label{sec:facecropresults}
We use LP3D's face cropping for our model, which includes the face and the shoulders. 
GPAvatar by default uses center crops (the largest square region at the center of an image) and do not perform face tracking. 
This could result in more or less complete reconstructions depending on the image.
Due to face cropping inconsistencies among the different methods, their numerical performance can vary based on the kind of cropping used for evaluation.
Our model also focuses on shoulders in addition to the head, thus we need to evaluate the models on different input/output image crops for fairness.

\textbf{"LP3D's Output Crop"} rows (Table. \ref{table:facecrop}): For reference, these numbers are also copied from the main paper.
Here each of the methods uses its default method to crop the inputs, but we crop the various methods' outputs using LP3D's cropping.

\textbf{"LP3D's Input Crop"} rows: These methods use the same cropped inputs as LP3D instead of applying LP3D's cropping to the output.

\textbf{"Face Crop"} rows: These methods use the same cropped inputs as LP3D, and the rendered images are later cropped around the face region using the face regions detected by the NVIDIA MAXINE AR SDK~\cite{maxineSDK} as on the groundtruth images. 
This cropping provides the most consistent cropping for all methods but also fails to measure important attributes like shoulder poses and hair.

Since the expression accuracy reported here was calculated using the NVIDIA MAXINE AR SDK~\cite{maxineSDK} on the face crop, the number reported here remains the same as in the main paper and across crops. The ArcFace identity loss is calculated on the entire image and thus varies slightly across different crops.

Our model is the best in expression and identity accuracy among all methods. 
Despite GPAvatar's good numerical performance on the LPIPS and PSNR metrics, its overall realism is significantly undermined by its dampened expression reconstruction, significant blurriness when viewed from the side, and often inaccurate reconstruction (Fig.~\ref{fig:comparison1}). 
Please refer to the supplementary video for more direct visual assessments.

\begin{table}[h]
\centering
\begin{tabularx}{\textwidth}{l | c | c | c | Y Y | Y Y}
\hline
 \multirow{2}{*}{Crop} & \multirow{2}{*}{Method} & \multirow{2}{*}{Expr$\downarrow$} & \multirow{2}{*}{ID$\downarrow$} & \multicolumn{2}{c|}{Overall Synthesis Quality} & \multicolumn{2}{c}{NVS Quality} \\
& & & & PSNR$\uparrow$ & LPIPS$\downarrow$ & PSNR$\uparrow$ & LPIPS$\downarrow$\\
\hline
Face & GPAvatar\cite{chu2024gpavatar} & 0.2041 & 0.2173  & 21.9434 & \textbf{0.2327} & \textbf{21.9434} & \textbf{0.2327} \\
Crop & LP3D\cite{trevithick2023} & \underline{0.1676} & \underline{0.1763} & \underline{21.5092} & 0.2511 & 20.7896  & 0.2670\\
& Ours & \textbf{0.1584} & \textbf{0.1644}  & \textbf{22.1388} & \underline{0.2494} & \underline{21.8849} & \underline{0.2546}\\
\hline
\multirow{1}{*}{LP3D's} & GPAvatar\cite{chu2024gpavatar} & 0.2041 & \underline{0.2026} & \underline{22.5624} & 0.2294 & \textbf{22.5624} & \textbf{0.2294} \\
\multirow{1}{*}{Input} & LP3D\cite{trevithick2023} & \underline{0.1676} & 0.2154 & 22.3309 & \underline{0.2232} & 21.5246  & 0.2374\\
\multirow{1}{*}{Crop} & Ours & \textbf{0.1584} & \textbf{0.1865}  & \textbf{22.7695} & \textbf{0.2189} & \underline{22.4395} & \underline{0.2240}\\
\hline
\multirow{1}{*}{LP3D's} & GPAvatar\cite{chu2024gpavatar} & 0.2041 & \underline{0.2074} & \underline{21.9487} & 0.2334 & 21.9487 & \underline{0.2334} \\
\multirow{1}{*}{Output} & LP3D\cite{trevithick2023} & \underline{0.1676} & 0.2154 & \underline{22.3309} & \underline{0.2232} & 21.5246  & 0.2374\\
\multirow{1}{*}{Crop} & \textbf{Ours} & \textbf{0.1584} & \textbf{0.1865}  & \textbf{22.7695} & \textbf{0.2189} & \textbf{22.4395} & \textbf{0.2240}\\
\hline
\end{tabularx}
\vspace{1pt}
\caption{\textbf{Comparison on NeRSemble~\cite{Nersemble} using face crops:} 
Quantitative performance on the NeRSemble~\cite{Nersemble} dataset using different input/output cropping settings.
The bottom \textbf{"LP3D's Output Crop"} rows: These numbers are included in the main paper, where each of the methods use their default methods to crop the inputs. 
We re-crop their outputs using LP3D's cropping method.
The middle \textbf{"LP3D's Input Crop"} rows: The methods use the same cropped inputs as LP3D instead of applying LP3D's cropping to their output.
The top \textbf{"Face"} rows: The methods use the same cropped inputs as LP3D, and the rendered images are cropped around the face region using NVIDIA MAXINE AR SDK's\cite{maxineSDK} detection. 
Our method achieves state-of-the-art expression and identity reconstruction across all cropping methods.
Please refer to the supplementary video for a better assessment of quality.
}
\label{table:facecrop}
\end{table}

\section{Joint vs. Separate Triplane Undistorter and Fuser}\label{sec:collapse}
\begin{figure}[t!]
  \centering
    \includegraphics[width=\textwidth]{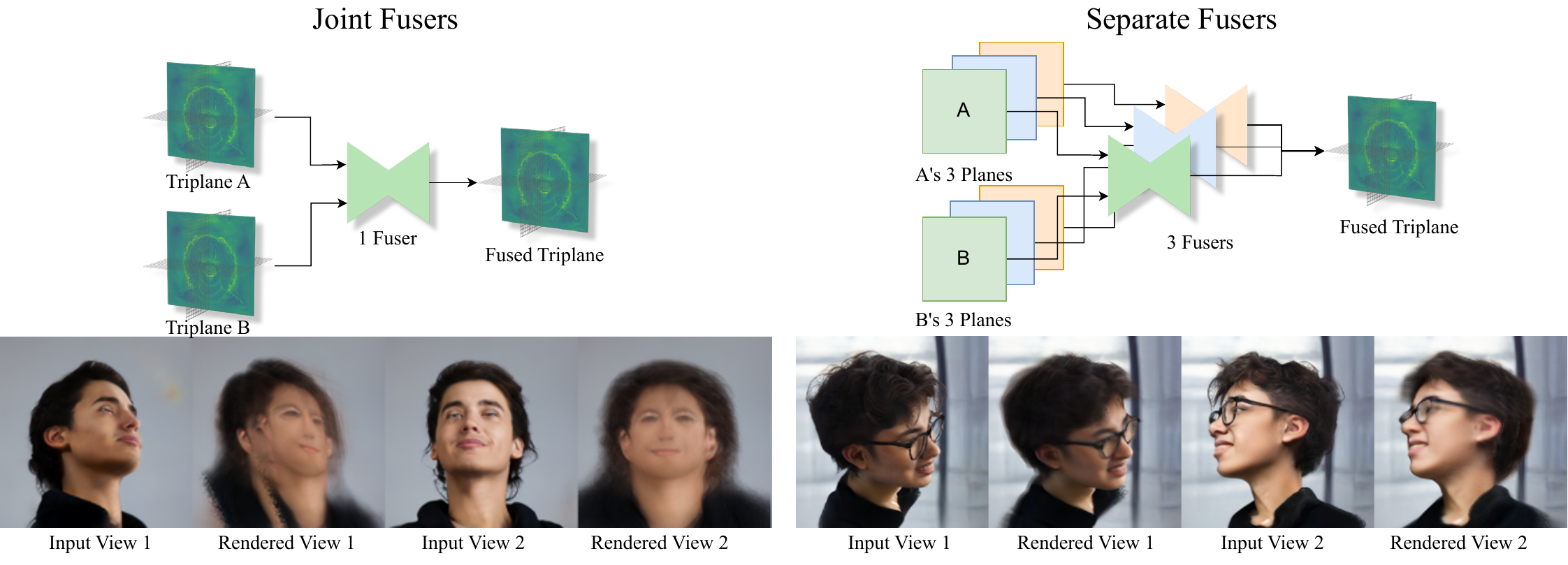}
   \caption{\textbf{Joint Fusion Causes Collapse to 2D:} \textit{Left}: The joint Fuser treats triplanes as a single feature image of size $96\times 256\times 256$. It uses one fusion network to combine two triplanes into a fused one. This approach leads to collapse to 2D as shown at the left-bottom. \textit{Right}: Using Separate Fusers effectively treats each plane in the triplane as a separate entity. Each of the 3 pairs of triplanes are fused separately and combined into the final fused triplane. This approach leads to correct fusion results.
   }
   \label{fig:split_fuser}
\end{figure}

As mentioned in the main paper, our Triplane Undistorter and Fuser modules both consist of 3 copies of the same network (with different weights), where each processes one plane in the triplane. 
It is not unexpected that the Undistorter would require three different instances for the three planes, because it estimates flow maps $T_{flow}\in \mathbb{R}^{3\times 2\times 256\times 256}$ using the optical flow architecture from SPyNet\cite{ranjan2017optical}. 
On the other hand, one might not expect that the Fuser to also require three different instances to process each plane separately. 
One might expect that jointly fusing the three planes using one transformer allows for communication of information between the 3 planes in a triplane and could thus improve results.
However, we find that using a single transformer leads to collapse to 2D (Fig.~\ref{fig:split_fuser} (left)). 
We also experimented with first projecting the feature planes into the same feature space before fusion. 
However, the results remain the same.
Whereas using 3 separate smaller networks results in correct fusion (Fig.~\ref{fig:split_fuser} (right)).
We suspect that this is because jointly fusing the triplanes is a significantly more difficult task than fusing each of the planes, separately.

\clearpage
\newpage
%
%

\end{document}